\documentclass[letterpaper, 10 pt, journal, twoside]{IEEEtran} 

\usepackage{times}
\usepackage{epsfig}
\usepackage{graphicx}
\usepackage{caption}
\usepackage{amsmath}
\usepackage{amssymb}
\usepackage{amsthm}
\usepackage{booktabs}
\usepackage{array}
\usepackage{subfig}
\usepackage{cite} % to group references
\usepackage{balance}
\usepackage{url}

\newcommand{\transpose}{\mathsf{T}}

\newtheorem{theorem}{Theorem}

\newtheorem{remark}{Remark}

\usepackage[ruled]{algorithm2e}
\newcommand{\nextnr}{\stepcounter{AlgoLine}\ShowLn}
\usepackage{algpseudocode}
\algnewcommand\algorithmicforeach{\textbf{for each}}
\algdef{S}[FOR]{ForEach}[1]{\algorithmicforeach\ #1\ \algorithmicdo}

\begin{document}

\title{A Closed-Form Uncertainty Propagation in Non-Rigid Structure from Motion}

\author{Jingwei Song, Mitesh Patel, Ashkan Jasour, and Maani Ghaffari%
	\thanks{J. Song and M. Ghaffari are with the University of Michigan, Ann Arbor, MI 48109, USA. \texttt{\{jingweso,maanigj\}@umich.edu}}%
	\thanks{M. Patel is with Nvidia Corporation, Santa Clara, CA - 95051, USA. \texttt{\{miteshp\}@nvidia.com}}%
	\thanks{A. Jasour is with Massachusetts Institute of Technology, Cambridge, MA 01239, USA \texttt{\{jasour\}@mit.edu}}
}

\maketitle

\begin{abstract}
Semi-Definite Programming (SDP) with low-rank prior has been widely applied in Non-Rigid Structure from Motion (NRSfM). A low-rank constraint avoids the inherent ambiguity of the basis number selection in conventional base-shape or base-trajectory methods. Despite  SDP-based NRSfM's efficiency, it remains unclear how to propagate the noisy tracked feature points' uncertainty to the 3D recovered shape in SDP-based NRSfM formulation. This paper presents a closed-form statistical inference for the element-wise uncertainty propagation of the estimated deforming 3D shape points in the exact low-rank SDP-based NRSfM. Then, we extend the exact low-rank uncertainty propagation to the approximate low-rank scenario with an optimal numerical rank selection method. The proposed method provides an independent module to the SDP-based method and only requires the statistical information of the input 2D trackings. Extensive experiments show that the major uncertainty in the recovered 3D points follows normal distribution, the proposed method quantifies the uncertainty accurately, and it has desirable effects on the routinely SDP low-rank based NRSfM solver. The code is open-sourced at \url{https://github.com/JingweiSong/NRSfM_uncertainty}.
\end{abstract}

\begin{IEEEkeywords}
NRSfM, uncertainty quantification, SDP 
\end{IEEEkeywords}

\IEEEpeerreviewmaketitle

%%%%%%%%% BODY TEXT
\section{Introduction}
\label{Section_introduction}

\IEEEPARstart{N}{on-Rigid} Structure from Motion (NRSfM) is the topic of recovering the camera motion and the 3D time-varying shape simultaneously from sequential 2D trajectories in the monocular video. It contributes to 3D shape perception in various applications, including 3D reconstruction and scene understanding, using consumer-level digital cameras.\par 

Uncertainty propagation has been heavily analyzed in robotics \cite{thrun2005probabilistic} and, Structure from Motion (SfM) \cite{irani2000factorization,morris1998unified} as the noise of the observation is not negligible. Uncertainty generally includes aleatoric uncertainty (relating to measurement noise), epistemic uncertainty (relating to model parameters), and structure uncertainty (relating to model structure) \cite{gal2016dropout}, but the uncertainty in SfM and Simultaneous Localization and Mapping (SLAM) is normally referred as aleatoric. The uncertainty propagation of the estimated state provides important statistical descriptions for further applications such as quality assessment \cite{andrew2001multiple}, localization \cite{kendall2016modelling}, mapping \cite{gan2020bayesian}, path planning \cite{van2011lqg}, and multi-source information fusion \cite{barfoot2014associating,du2019multiresolution}. In NRSfM community, \cite{agudo2015sequential,agudo2021total} propagate uncertainty sequentially with their modified Extended Kalman Filter (EKF), which is coupled with map deformation modeling. We do not find any uncertainty estimation algorithm for the widely used Semi-Definite Programming (SDP) based NRSfM. The lack of uncertainty analysis in SDP-based NRSfM may be attributed to the difficulty in linearizing the objective function or the discontinuity of the objective function. Unlike in robotics and SfM, where the sensor-to-object functions can be linearized easily, NRSfM involves complicated nonlinear formulation such as factorization or nuclear norm minimization, which hinders the process of uncertainty propagation. 

\begin{figure}[t]
    \centering
    \includegraphics[width=0.99\columnwidth]{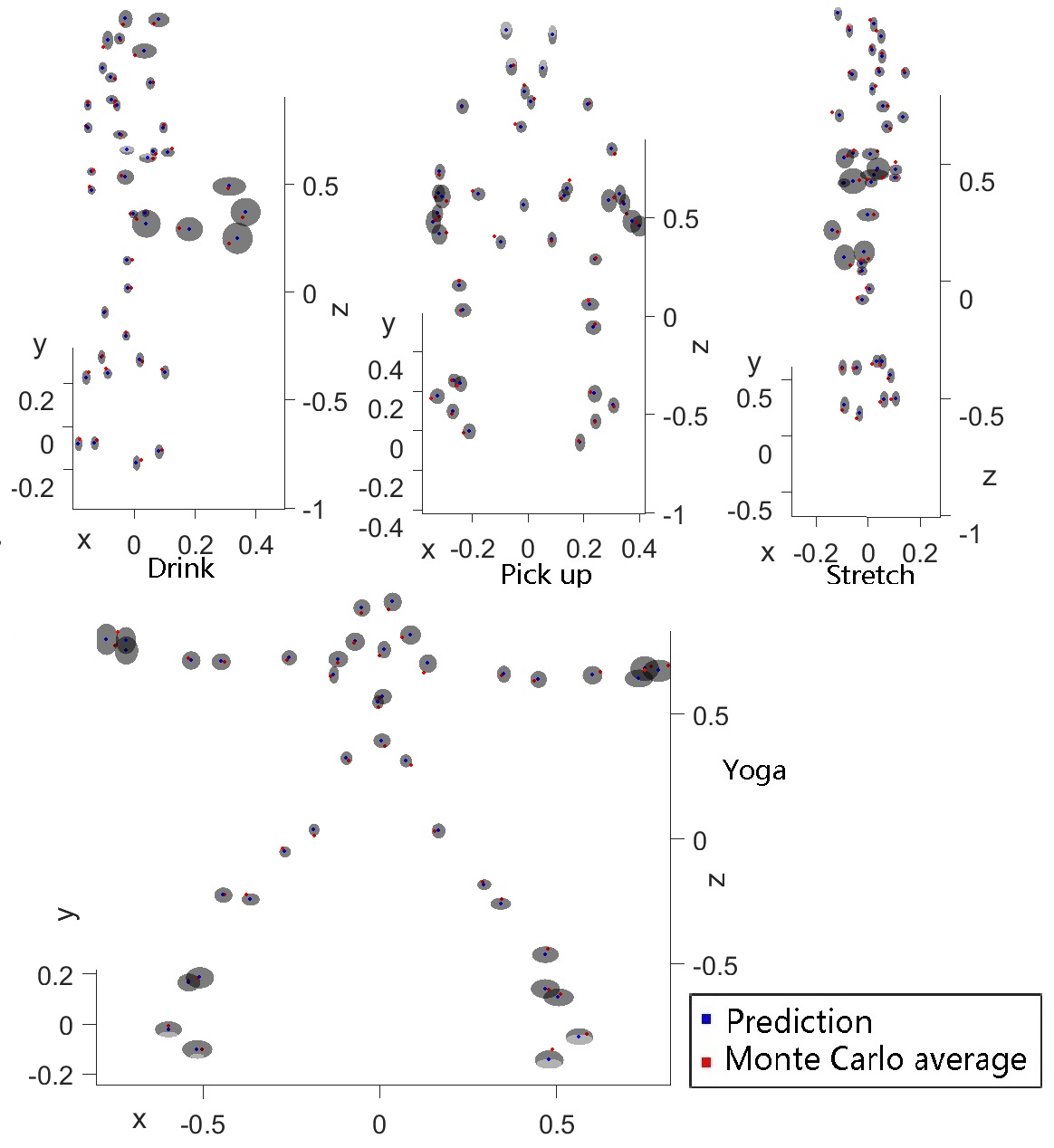}
    \caption{Presented is the sample reconstructions (in blue), average of 100 Monte Carlo results (in red) and error ellipses of the recovered points. The error ellipse denotes 1.96 sigma (observation noise sigma is $0.05$) bound. Readers are encouraged to watch the attached video for more information.}
    \label{fig_error_ellipse}
\end{figure}

%This research aims to overcome the difficulty of uncertainty inference from non-linear NRSfM formulation by proposing a general uncertainty propagation for both factorization and nuclear norm formulations. 

%   缩写版
%This work proposes a closed-form solution to the element-wise uncertainty propagation for the group of Semi-Definite Programming (SDP) based NRSfM \cite{dai2014simple}. To be consistent with the rank-free prior, we propose an empirical approximate rank estimation method and show that rank has a remarkably small impact on covariance matrix estimation. Novelties are: 1. This is the first research addressing the closed-form element-wise uncertainty propagation algorithm for the NRSfM problem; 2. An algorithm is presented to obtain the optimal rank for the approximate time-varying shape structure in noisy environment.\par

This work proposes a closed-form solution to the element-wise uncertainty propagation for the group of SDP-based NRSfM \cite{dai2014simple} formulations. Extensive experiments show that the proposed uncertainty estimation algorithm describes the estimated shape well and can be applied in measuring confidences of the recovered 3D deformable shape. Additionally, we adapt the exact rank algorithm with an empirical approximate rank estimation method for the prior-free SDP-based approaches. Results show that the rank has a remarkably small impact on the element-wise variance estimation. 

It should be noticed that we only aim at uncertainty estimation of the recovered deforming shape. Uncertainty of the camera rotation matrix cannot be obtained due to ambiguity. \cite{akhter2009defense} shows that the camera rotation matrix and estimated shape cannot be retrieved simultaneously due to their ambiguity. Normally the camera pose is estimated with other off-the-shelf techniques or rigid background \cite{dai2014simple,agudo2017dust}. Thus, our work only focuses on the 3D shape uncertainty in SDP-based NRSfM. In particular, this work has the following contributions.
\begin{enumerate}
    \item We propose a closed-form element-wise uncertainty propagation algorithm for SDP-based NRSfM.
    \item A numerical rank estimation method is proposed to define the best rank for the confidence estimation algorithm.
    \item The code is publicly available for future research and benchmark comparison. \footnote{Code is available at \url{https://github.com/JingweiSong/NRSfM_uncertainty.git}.}
\end{enumerate}

% \begin{itemize}
%     \item This is the first research addressing the closed-form element-wise uncertainty propagation algorithm for the SDP-based NRSfM. 
%     \item A numerical rank estimation method is presented to define the best rank for the confidence interval. 
% \end{itemize}

\section{Related Work}
This section briefly reviews the state-of-the-art works in NRSfM and rigid uncertainty propagation methods.
\subsection{NRSfM}
% \textbf{Review of NRSfM}. 
Following the rigid SfM community, early NRSfM works \cite{bregler2000recovering} directly adopted factorization to recover the time-varying 3D shapes. Later, \cite{xiao2004closed} showed ambiguous solvability in NRSfM that factorization alone is insufficient in solving the ill-posed NRSfM problem. Since then, priors have been introduced to constrain the problem into a low-rank subspace to ensure solvability. The low-rank priors include base shape \cite{bartoli2008coarse,torresani2008nonrigid}, base trajectory \cite{akhter2009nonrigid,valmadre2012general,bartoli2008coarse}, base shape-trajectory \cite{gotardo2011kernel,gotardo2011computing,simon2014separable} and force model \cite{agudo2018force}. There is a duality in the formulation of all these proposed base prior, which is, the estimated state (shape or trajectory) is a linear combination of all bases. Innovations include `coarse-to-fine' \cite{bartoli2008coarse}, probabilistic principal component analysis \cite{torresani2008nonrigid}, kernel trick \cite{gotardo2011kernel}, and Procrustean analysis in consecutive shapes \cite{lee2014procrustean} were proposed. As \cite{akhter2009defense} pointed out, the internal constraint orthonormality of camera orientation and external constraint of fixed base shapes enable the unique shape structure. \par 

One milestone was achieved by \cite{fragkiadaki2014grouping} and \cite{dai2014simple} who enforced low-rank constraints for spatial-temporal smoothness of the shape. Different from conventional basis formulation, which is in a hard low-rank constraint, the SDP-based formulation is in the soft low-rank loss (nuclear norm for the objective function's convexity) minimization. As \cite{dai2014simple} pointed out, the number of basis is not essential in their SDP formulation and thus can be classified as prior-free. The only constraint needed is the low nuclear norm constraint in their SDP-based formulation. The automatic prior-free SDP achieves high accuracy with fewer parameter configurations. SDP and their derived methods are the most widely used algorithms in the state-of-the-art NRSfM community. Following researchers pushed the low-rank formulation toward spanning the model with a union of low dimensional shape subspace \cite{cabral2013unifying,zhu2014complex,agudo2017dust,kumar2017spatio,gu2018monocular,Golyanik_2019,kumar2019jumping,kumar2020non,kumar2020dense,parashar2021robust,agudo2020unsupervised} for multiple body reconstruction. \cite{Golyanik2020DSPR} successfully filled the gap between the theory and application. Other prior-based learning-based methods include \cite{kong2016prior,kong2019deep,wang2019distill,sidhu2020neural,park2020procrustean}. These `training and testing' methods fall out of the scope of this article. This study concentrates on the prior-free and training-free SDP-based methods.\par

% To relax the prior-free requirement to `training and testing', learning-based methods \cite{kong2016prior,kong2019deep,wang2019distill,sidhu2020neural,park2020procrustean} are introduced recently in this geometric problem which claims to enable interpretable deformable 3D shape recovery. Different from the above methods, these prior-based approaches are in the `training and testing' procedure, and a sparse dictionary is pre-trained to encode prior knowledge for representing 3D geometry. The trained dictionary is adopted to build the time-varying shapes. In the `training and testing' procedure, the prediction is closely related to the training step and the similarity between the training and testing data sets, thus the uncertainty is related to the two factors just as mutual coherence mentioned in \cite{kong2019deep}. The `training and testing' methods fall out of the scope of this paper, and we only focus on the prior-free and training-free low-rank methods.\par 

\subsection{Rigid Uncertainty Propagation}
The uncertainty propagation has been extensively studied in SfM, and SLAM communities. The confidences are of equal importance as the predictions \cite{hartley2003multiple}. In SfM or SLAM, uncertainty is defined as the impact of input perturbation and measured as covariance matrices. SfM process normally linearizes the objective function and solves with Gauss-Newton algorithm \cite{morris1999uncertainty}, which defines the covariance as the inverse of the second-order derivations known as `Hessian matrix'. Numerically, the Hessian matrix is approximated by $\mathbf{H}=\mathbf{J}^\transpose\mathbf{J}$ where $\mathbf{J}$ is the first-order derivative. In sequential state estimation, SLAM \cite{bailey2006consistency} propagates the uncertainty through the linearized state propagation function in the EKF. With known initial pose and map uncertainties, SfM system balances the weights to yield the optimal estimation and the associated uncertainty. Built on the EKF theory and deformation modeling, the sequential NRSfM \cite{agudo2015sequential,agudo2021total} algorithms, or termed deformable SLAM, achieved camera orientation and map uncertainty propagation iteratively in the dynamic process.\par 

The closed-form uncertainty estimation in SDP-based NRSfM, however, is far from the straightforward solutions. Although numerous researches are conducted in the rigid scenario or sequential NRSfM, no work analyzes its related uncertainty. Due to the nonlinear nuclear norm constraint, SDP fails to pin the formulation down to a closed-form version as SLAM or SfM. The nuclear norm term is an obstruct for closed-form uncertainty propagation. Recently, a breakthrough has been achieved in optimal uncertainty propagation and inference for noisy matrix completion \cite{chen2019inference}. The relaxed convex optimization in matrix completion shares the similarity in SDP in NRSfM \cite{dai2014simple}. It measures the confidence interval by developing simple de-biased estimators that admit tractable and accurate distributional characterizations. This article is inspired by \cite{chen2019inference} and provides a solution to quantify the uncertainty of the estimated time-varying shape. Moreover, \cite{chen2019inference} only presented a solution to the matrix with an exact low-rank structure. We provide a numerical solution in NRSfM to allow uncertainty propagation for the approximate low-rank matrix form.\par

\section{Methodology}
\subsection{Problem definition and the SDP solver}
The classic SDP-based NRSfM \cite{dai2014simple} formulates the time-varying shape recovery as minimizing
\begin{equation}
\label{Eq_NRSfM_objective}
\begin{array}{l}
{\underset{\mathbf{S},\mathbf{R}}{\min} \  \mu\lVert{\mathbf{S}^\sharp}\lVert_{*}+\frac{1}{2}\lVert\mathbf{W}-\mathbf{R}\mathbf{S}\lVert_\mathrm{F}^{2}, \text {\  such that }} \\ {\mathbf{S}^{\sharp}=g(\mathbf{S})=[\mathbf{P}_{X} \mathbf{P}_{Y} \mathbf{P}_{Z}](\mathbf{I}_{3} \otimes \mathbf{S}) \text {\ and}}\\
{ \mathbf{R}\mathbf{R}^{\transpose}=\mathbf{I}_{2F}}
\end{array},
\end{equation}

% \begin{align}
% \label{Eq_SDefinition}
% %\arraycolsep=3pt\def\arraystretch{1.8}
% {\mathbf{S}}=
% \left(
% \begin{array}{ccccccccc}
% x_1^1&y_1^1&z_1^1&\cdots&x_1^F&y_1^F&z_1^F\\
% \vdots&\vdots&\vdots&\vdots&\vdots&\vdots&\vdots\\
% x_N^1&y_N^1&z_N^1&\cdots&x_N^F&y_N^F&z_1^F\\
% \end{array}
% \right)^\transpose,
% \end{align}

% \begin{align}
% \label{Eq_S_sharpDefinition}
% %\arraycolsep=3pt\def\arraystretch{1.8}
% {\mathbf{S}^{\sharp\mathrm{d}}}=
% \left(
% \begin{array}{ccccccccc}
% x_1^1&\cdots&x_N^1&y_1^1&\cdots&y_N^1&z_1^1&\cdots&z_N^1\\
% \vdots&\ddots&\vdots&\vdots&\ddots&\vdots&\vdots&\ddots&\vdots\\
% x_1^F&\cdots&x_N^F&y_1^F&\cdots&y_N^F&z_1^F&\cdots&z_N^F\\
% \end{array}
% \right)^\transpose,
% \end{align}
% \noindent where the time-varying shape ${\mathbf{S}}$ ( \eqref{Eq_SDefinition}) is a $3F \times N$ matrix with $F$
\noindent where $\lVert\cdot\lVert_*$ is the nuclear norm, $\lVert\cdot\lVert_\mathrm{F}$ is the Frobenius norm and $\otimes$ is the Kronecker product. $\mathbf{I}_{2F}$ is the identity matrix with size $2F$. $\mu$ is the hyper-parameter balancing two constraints. The time-varying shape ${\mathbf{S}}$ %( \eqref{Eq_SDefinition}) 
is a $3F \times N$ matrix with $F$ frames and $N$ 3D feature points ($\mathbf{p}^i_j=[x_i^j\ y_i^j\ z_i^j]$ point $j$ in frame $i$) which are sequentially permuted. 
%Eq. \eqref{Eq_S_sharpDefinition} shows the definition of 
${\mathbf{S}^\sharp}$ is the repermuted matrix of ${\mathbf{S}}$ and in the size $3N \times F$. It is the rearranged form of the matrix ${\mathbf{S}}$ for conciseness.  $\mathbf{S}^{\sharp}=g(\mathbf{S})$ maps $\mathbf{S}$ to $\mathbf{S}^{\sharp}$ while ${\mathbf{S}}=g^{-1}(\mathbf{S}^{\sharp})$ does the opposite. $\mathbf{P}_{X}, \mathbf{P}_{Y}, \mathbf{P}_{Z} \in \mathbb{R}^{F \times 3 F}$ are some properly defined 0-1-valued `row-selection' matrices (similar to the `permutation matrix'). $\mathbf{R}$ is the camera rotation matrix composed of the diagonal block matrix $\mathbf{R}_i$. $\mathbf{R}_i$ is the first two rows of its full rotation form $\tilde{\mathbf{R}}_i \in SO(3)$. $\mathbf{R}$ defines the 3D to 2D  projection. As \cite{agudo2017dust} pointed out, there are several off-the-shelf methods to estimate the camera rotation matrix $\mathbf{R}$ prior to \eqref{Eq_NRSfM_objective}. Thus, ${\mathbf{R}}$ is estimated firstly and assumed noiseless. $\mathbf{W} \in \mathbb{R}^{2F \times N}$ is the noisy observation matrix of the 2D trackings as

\begin{equation}
\label{Eq_observationGaussian}
\begin{split}
\mathbf{W}_{i j}={\mathbf{W}}_{i j}^{\star}+\mathbf{E}_{i j},  \mathbf{E}_{i j} \stackrel{\mathrm{i.i.d.}}{\sim} \mathcal{N}(0, \sigma_0^{2}), 
  \text {for all}(i,j) \in \Omega,
\end{split}
\end{equation}
\noindent where $\Omega \subseteq\{1, \cdots, 2F\} \times\{1, \cdots, N\}$ is the subset of indexes, ${\mathbf{W}^{\star}_{ij}}$ is the noise-free observation at point $j$ in frame $i$. The uniform Gaussian noise $\mathbf{E}_{i j}$ denotes the spatial disturbance of element $(i j)$ in measurement. $\sigma_0^{2}$ is the variance of the Gaussian noise. The permuted version $\mathbf{E}^\sharp = g(\mathbf{E})$.

% \begin{align}
% \label{NRSfM_form}
% {\mathbf{W}}
% ={\mathbf{R}}{\mathbf{S}}.
% \end{align}

% \begin{align}
% \label{NRSfM_form}
% \arraycolsep=3pt\def\arraystretch{1.8}
% \underbrace{
% 	\left(
% 	\begin{array}{ccc}
% 	{\mathbf{u}}^1_1&\cdots&{\mathbf{u}}^N_1\\
% 	\vdots&\ddots&\vdots\\
% 	{\mathbf{u}}^1_F&\cdots&{\mathbf{u}}^N_F\\
% 	\end{array}
% 	\right)
% }_{\mathbf{W}}
% =
% \underbrace{
% 	\left(
% 	\begin{array}{ccc}
% 	\mathbf{R}_1&&\\
% 	&\ddots&\\
% 	&&\mathbf{R}_F\\
% 	\end{array}
% 	\right)
% }_{\mathbf{R}}
% \underbrace{
% 	\left(
% 	\begin{array}{ccc}
% 	\mathbf{p}^1_1&\cdots&\mathbf{p}^N_1\\
% 	\vdots&\ddots&\vdots\\
% 	\mathbf{p}^1_F&\cdots&\mathbf{p}^N_F\\
% 	\end{array}
% 	\right)
% }_{\mathbf{S}}.
% \end{align}

%By imposing the orthographic constraint, the conventional NRSfM method adopts rank $3K$ factorization \cite{vidal2006nonrigid,bregler2000recovering}, where $K$ is deliberately chosen, and results are highly dependent on the choice of $K$.\par 

Equation \eqref{Eq_NRSfM_objective} was proposed to overcome the difficulty in selecting the optimal rank in searching for $\mathbf{S}$ \cite{dai2014simple}. The low-rank constraint makes the objective function nonconvex. Therefore, nuclear norm minimization is adopted to relax the problem to convex form \cite{dai2014simple,fragkiadaki2014grouping}. \par

\subsection{Uncertainty propagation in the low-rank structure}
\label{section_exact_structure}

\begin{figure}[t]
	\centering
	\subfloat{
% 		\begin{minipage}[]{0.5\textwidth}
% 			\centering
			\includegraphics[width=1\columnwidth]{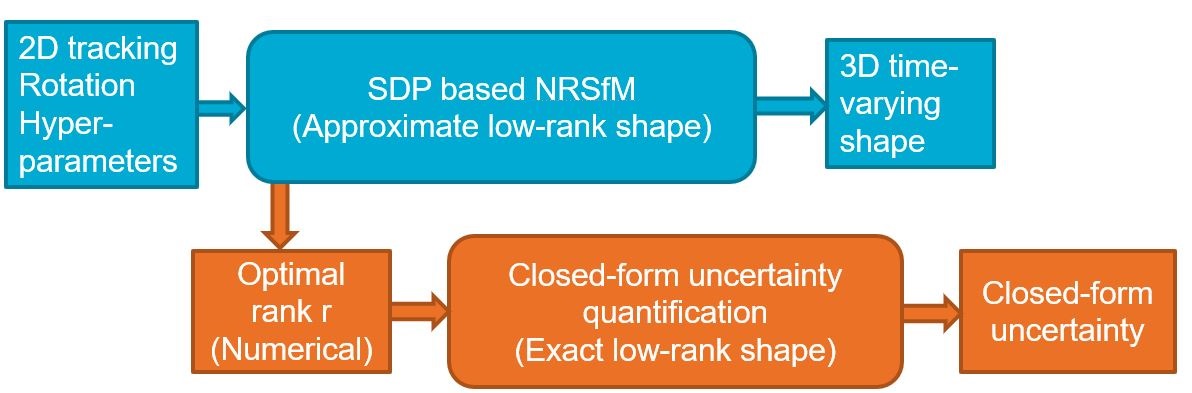}
% 		\end{minipage}
	}
	\caption{The diagram shows the relation between the proposed method and the previous research. The blocks in blue are the process of SDP-based NRSfM \cite{dai2014simple} while the blocks in orange are the process of our approach.}
	\label{fig:Diagram_everyone}
\end{figure}

Fig. \ref{fig:Diagram_everyone} presents the relationship of the proposed method and previous research \cite{dai2014simple}. Our work proposes a new module (in orange) to quantify the uncertainty of the SDP-based deformable 3D shape estimation (in blue). This section presents a method to retrieve the closed-form uncertainty in the scenario that the time-varying shape is in the exact low-rank. Section \ref{subsection_optimal_rank} extends it to the approximate low-rank version. Note that in the SDP formulation \eqref{Eq_NRSfM_objective}, the ground truth of the time-varying deforming shape is in approximate low-rank structure, meaning it is strictly full-rank but has a small nuclear norm, which can be viewed as the approximated low-rank structure. It should be noticed that the impact of noise on the rotation matrix $\mathbf{R}$ is not considered because it is obtained separately.\par

%But this section only addresses the exact low-rank case.\par

% In this section, we will show that the closed-form uncertainty propagation for the exact low-rank structure can be retrieved with mathematical manipulation and approximation. After the achievement of the closed-form uncertainty propagation of the exact low-rank form, a practical approach is presented to extend it to the approximate version. \par 

Denote $\hat{\mathbf{S}}^{\sharp} \in \mathbb{R}^{3N \times F}$ as the  rank-$r$ matrix estimated from the noise-free observation $\mathbf{W}^{\star}$ (the exact low-rank structure case) after solving \eqref{Eq_NRSfM_objective}. Please note that $\hat{\mathbf{S}}^{\sharp}$ is \emph{not} the ground truth but the noise-free estimation.
Denote $\mathbf{S}^{\sharp\mathrm{d}} \in \mathbb{R}^{3N \times F}$ as the rank-$r$ matrix estimated from the noisy observation $\mathbf{W}$. Enforce Singular Value Decomposition (SVD) and we have $\hat{\mathbf{S}}^{\sharp}=\hat{\mathbf{U}} \hat{\mathbf{\Sigma}} \hat{\mathbf{V}}^{\transpose}$ and $\mathbf{S}^{\sharp\mathrm{d}}=\mathbf{U}^{\mathrm{d}} \mathbf{\Sigma}^{\mathrm{d}} \mathbf{V}^{\mathrm{d} \transpose} $. We define two auxiliary matrices $\hat{\mathbf{X}} := \hat{\mathbf{U}} \hat{\mathbf{\Sigma}}^{ 1 / 2} \in \mathbb{R}^{3N\times r}$ and $\hat{\mathbf{Y}} := \hat{\mathbf{V}} \hat{\mathbf{\Sigma}}^{ 1 / 2} \in \mathbb{R}^{F \times r}$, where $\hat{\mathbf{X}}$ and $\hat{\mathbf{Y}}$ are the globally rectified matrices. Similarly, define $\mathbf{X}^{\mathrm{d}} \in \mathbb{R}^{3N \times r}, \mathbf{Y}^{\mathrm{d}} \in \mathbb{R}^{F \times r}$ as unrectified estimations. The following rules apply.\par 

% \begin{equation}
% \begin{aligned}
% \hat{\mathbf{X}}^{\transpose} \hat{\mathbf{X}}=\hat{\mathbf{Y}}^{\transpose} \hat{\mathbf{Y}}=\hat{\mathbf{\Sigma}}\\
% \hat{\mathbf{S}}^{\sharp}=\hat{\mathbf{X}}{\hat{\mathbf{Y}}^{\transpose}}.
% \end{aligned}
% \end{equation}

\begin{align}
%\label{Eq_relation_X_Sigma}
\begin{array}{l}\hat{\mathbf{X}}^{\transpose} \hat{\mathbf{X}}=\hat{\mathbf{Y}}^{\transpose} \hat{\mathbf{Y}}=\hat{\mathbf{\Sigma}},\ 
	\hat{\mathbf{S}}^{\sharp}=\hat{\mathbf{X}}{\hat{\mathbf{Y}}^{\transpose}} \\ 
	\mathbf{X}^{\mathrm{d} \transpose} \mathbf{X}^{\mathrm{d}}=\mathbf{Y}^{\mathrm{d} \transpose} \mathbf{Y}^{\mathrm{d}}=\mathbf{\Sigma}^{\mathrm{d}},\ 
	\mathbf{S}^{\sharp\mathrm{d}}=\mathbf{X}^{\mathrm{d}}{\mathbf{Y}^{\mathrm{d}\transpose}}\end{array}.
\end{align}

%
%\begin{align}
%\label{Eq_relation_X_Sigma}
%% \begin{aligned}
%\hat{\mathbf{X}}^{\transpose} \hat{\mathbf{X}}=\hat{\mathbf{Y}}^{\transpose} \hat{\mathbf{Y}}=\hat{\mathbf{\Sigma}},\ 
%\hat{\mathbf{S}}^{\sharp}=\hat{\mathbf{X}}{\hat{\mathbf{Y}}^{\transpose}},
%% \end{aligned}
%\end{align}
%
%\begin{align}
%\label{Eq_relation_X_Sigma_1}
%% \begin{aligned}
%\mathbf{X}^{\mathrm{d} \transpose} \mathbf{X}^{\mathrm{d}}=\mathbf{Y}^{\mathrm{d} \transpose} \mathbf{Y}^{\mathrm{d}}=\mathbf{\Sigma}^{\mathrm{d}},\ 
%\mathbf{S}^{\sharp\mathrm{d}}=\mathbf{X}^{\mathrm{d}}{\mathbf{Y}^{\mathrm{d}\transpose}}.
%% \end{aligned}
%\end{align}

In conventional NRSfM in shape basis formulation, $\hat{\mathbf{Y}}$ is termed as the basis matrix and $\hat{\mathbf{X}}$ is the linear coefficient matrix  \cite{akhter2009nonrigid}. Following the ambiguity issue in the coefficient matrix and basis matrix raised by \cite{akhter2009defense}, an optimal global rectification matrix $\mathbf{H}^{\mathrm{d}} \in \mathbb{R}^{r \times r}$ is needed to align the noisy estimation ($\mathbf{X}^{\mathrm{d}}$, $\mathbf{Y}^{\mathrm{d}}$) to the noise-free and rectified estimation ($\hat{\mathbf{X}}$, $\hat{\mathbf{Y}}$). $\mathbf{H}^{\mathrm{d}}$ can be obtained as

\begin{align}
\label{Eq_rectify}
\mathbf{H}^{\mathrm{d}}\! :=\! \arg\! \min _{\mathbf{H} \in \mathcal{A}^{r \times r}}\lVert\mathbf{X}^{\mathrm{d}} \mathbf{H}\!-\!\hat{\mathbf{X}}\lVert_{\mathrm{F}}^{2}\!+\!\lVert\mathbf{Y}^{\mathrm{d}} \mathbf{H} \!-\!\hat{\mathbf{Y}}\lVert_{\mathrm{F}}^{2},
\end{align}
\noindent where $\mathcal{A}^{r \times r}$ denotes orthogonal matrix manifold.\par

\begin{theorem}[]
\label{theo_main}
On condition that the observation noise matrix $\mathbf{E}$ is i.i.d., follows Gaussian and is relatively small, the errors between the estimated operator $(\mathbf{X}^{\mathrm{d}},\mathbf{Y}^{\mathrm{d}})$ from the noisy observation $\mathbf{W}$ and the estimated operator $(\hat{\mathbf{X}},\hat{\mathbf{Y}})$ from the noise-free observation $\mathbf{W}^{\star}$ are
\begin{align}
\label{Eq_ZxZyDefine}
\begin{array}{l}{\mathbf{X}^{\mathrm{d}}\mathbf{H}^{\mathrm{d}}-\hat{\mathbf{X}}=\mathbf{Z}_X}+\mathbf{\Psi}_\mathrm{X} \\ {\mathbf{Y}^{\mathrm{d}} \mathbf{H}^{\mathrm{d}}-\hat{\mathbf{Y}} =\mathbf{Z}_Y}+\mathbf{\Psi}_\mathrm{Y}\end{array},
\end{align}
\noindent where $\mathbf{\Psi}_\mathrm{X}  \in \mathbb{R}^{3N \times r}$ and $\mathbf{\Psi}_\mathrm{Y} \in \mathbb{R}^{F \times r}$ are the non-Gaussian residual matrices. $\mathbf{\Psi}_\mathrm{X}$ and $\mathbf{\Psi}_\mathrm{Y}$ are always smaller than Gaussian error matrices $\mathbf{Z}_X$ and $\mathbf{Z}_Y$. The rows of the error matrix $\mathbf{Z}_X \in \mathbb{R}^{3N \times r}$ (resp. $\mathbf{Z}_Y \in \mathbb{R}^{F \times r}$) are independent and obey
\begin{align}
\label{Eq_keyconclusion}
\begin{array}{ll}{\mathbf{Z}_X^{\transpose} \mathbf{e}_j \stackrel{\text { i.i.d. }}{\sim} \mathcal{N}(0, \frac{3}{2}\sigma_0^{2}(\mathbf{\Sigma}^{\mathrm{d}})^{-1}),} & {\text { for } 1 \leq j \leq r} \\ {\mathbf{Z}_Y^{\transpose} \mathbf{e}_j \stackrel{\text { i.i.d. }}{\sim} \mathcal{N}(0, \frac{3}{2}\sigma_0^{2}(\mathbf{\Sigma}^{\mathrm{d}})^{-1}),} & {\text { for } 1 \leq j \leq r}\end{array},
\end{align}
\noindent where $\mathbf{e}_j$ is the base vector with one element $1$ and rest $0$. \eqref{Eq_keyconclusion} strictly holds on condition that the frame size $F$ (observed from different orientations) is infinite. Appendix proves Theorem \ref{theo_main}. \par
\end{theorem}

%\textbf{Remark 1}. Unless addressed, the approximation in this paper is made assuming that the solution converges close to the global minimum. That is $\mathbf{\Psi_X}$ and $\mathbf{\Psi_Y}$ are significantly smaller than $\mathbf{Z}_X$ and $\mathbf{Z}_Y$. The closer it converges, the more reliable is the uncertainty propagation. This definition is similar to the uncertainty from the Bayesian estimation which quantifies the confidence of the average prediction.\par
%
%When algorithm converges, \eqref{Eq_ZxZyDefine} is approximated to: \par 
%
%\begin{align}
%\label{Eq_ZxZyDefine_1}
%\begin{array}{l}{\mathbf{X}^{\mathrm{d}} \mathbf{H}^{\mathrm{d}}-\mathbf{X}^{\star}\approx\mathbf{Z}_X} \\ {\mathbf{Y}^{\mathrm{d}} \mathbf{H}^{\mathrm{d}}-\mathbf{Y}^{\star}\approx\mathbf{Z}_Y}\end{array}.
%\end{align}

%Assuming that the first-order expansion is reasonably tight, 

Based on Theorem \ref{theo_main}, the element-wise difference between the estimation $\mathbf{S}_{i j}^{\sharp\mathrm{d}}$ from the noisy observation and the estimation $\hat{\mathbf{S}}_{i j}^{\sharp}$ from the noise-free observation is

\begin{equation}
%\label{Eq_error_S_Ssharp}
\begin{aligned} 
&\mathbf{S}_{i j}^{\sharp\mathrm{d}}-\hat{\mathbf{S}}_{i j}^{\sharp} 
=[\mathbf{X}^{\mathrm{d}} \mathbf{H}^{\mathrm{d}}(\mathbf{Y}^{\mathrm{d}} \mathbf{H}^{\mathrm{d}})^\transpose-\hat{\mathbf{X}} \hat{\mathbf{Y}}^{\transpose}]_{i j}\\
&=[\mathbf{e}_{i}^{\transpose}\mathbf{X}^{\mathrm{d}}\mathbf{H}^{\mathrm{d}}(\mathbf{Y}^{\mathrm{d}} \mathbf{H}^{\mathrm{d}})^\transpose\mathbf{e}_{j}-\mathbf{e}_{i}^{\transpose}(\hat{\mathbf{X}} {\hat{\mathbf{Y}}}^\transpose) \mathbf{e}_{j}]\\
&=[\mathbf{e}_{i}^{\transpose}\mathbf{X}^{\mathrm{d}}\mathbf{H}^{\mathrm{d}}(\mathbf{Y}^{\mathrm{d}} \mathbf{H}^{\mathrm{d}})^\transpose\mathbf{e}_{j}-\\
&\mathbf{e}_{i}^{\transpose}(\mathbf{X}^{\mathrm{d}}\mathbf{H}^{\mathrm{d}}-\mathbf{Z}_X)( \mathbf{Y}^{\mathrm{d}} \mathbf{H}^{\mathrm{d}}-\mathbf{Z}_Y)^\transpose) \mathbf{e}_{j}]\\
%&=[\mathbf{e}_{i}^{\transpose}(\mathbf{X}^{\mathrm{d}} \mathbf{H}^{\mathrm{d}}-\hat{\mathbf{X}}) \hat{\mathbf{Y}}^{\transpose} \mathbf{e}_{j}+\mathbf{e}_{i}^{\transpose} \hat{\mathbf{X}}(\mathbf{Y}^{\mathrm{d}} \mathbf{H}^{\mathrm{d}}-\hat{\mathbf{Y}}^{\transpose}) \mathbf{e}_{j}] \\ 
& \stackrel{(\mathrm{i})}\approx [\mathbf{e}_{i}^{\transpose} (\mathbf{Z}_X (\mathbf{Y}^{\mathrm{d}} \mathbf{H}^{\mathrm{d}})^\transpose) \mathbf{e}_{j}+\mathbf{e}_{i}^{\transpose} (\mathbf{X}^{\mathrm{d}}\mathbf{H}^{\mathrm{d}} \mathbf{Z}_Y^{\transpose}) \mathbf{e}_{j}], 
\end{aligned}
\end{equation}

\noindent where the bases $\mathbf{e}_i,\ \mathbf{e}_j$ localize the elements involved in calculating the elements in location $(i,j)$. The higher order error term $\mathbf{e}_{i}^{\transpose}\mathbf{Z}_X \mathbf{Z}_Y^ \transpose \mathbf{e}_{j}$ is neglected in $(\mathrm{i})$. 
%$[\cdot]_{ij}$ is the $(i,j)$ element of $\cdot$. 
After some manipulation, we have the element-wise variance of the error as 
% \begin{align}
\begin{equation}
\begin{aligned}
&{\operatorname{Var}}(\mathbf{S}_{i j}^{\sharp\mathrm{d}}-\hat{\mathbf{S}}_{i j}^{\sharp})
\stackrel{(\mathrm{i})}{=} [{\operatorname{Var}}(\mathbf{e}_{i}^{\transpose} (\mathbf{Z}_X {(\mathbf{Y}^{\mathrm{d}} \mathbf{H}^{\mathrm{d}})^\transpose}) \mathbf{e}_{j})+\\
&{\operatorname{Var}}(\mathbf{e}_{i}^{\transpose} ((\mathbf{X}^{\mathrm{d}} \mathbf{H}^{\mathrm{d}}) \mathbf{Z}_Y^{\transpose}) \mathbf{e}_{j})] \\
&\stackrel{(\text { ii })}{=} \frac{3}{2}\sigma_0^{2}[{\mathbf{e}_{j}}^{\transpose} (\mathbf{Y}^{\mathrm{d}} \mathbf{H}^{\mathrm{d}})(\mathbf{\Sigma}^{\mathrm{d}})^{-1} {(\mathbf{Y}^{\mathrm{d}} \mathbf{H}^{\mathrm{d}})^\transpose} {\mathbf{e}_{j}}+\\
&\mathbf{e}_{i}^{\transpose}\! (\mathbf{X}^{\mathrm{d}} \mathbf{H}^{\mathrm{d}})\!(\mathbf{\Sigma}^{\mathrm{d}})^{-1}\! (\mathbf{X}^{\mathrm{d}} \mathbf{H}^{\mathrm{d}})^\transpose\! \mathbf{e}_{i}]
\!\stackrel{(\text{iii})}{=}\! \frac{3}{2}\sigma_0^2(\lVert\mathbf{U}_{i,.}^{\mathrm{d}}\lVert_{2}^{2}\!+\!\lVert\mathbf{V}_{j,.}^{\mathrm{d}}\lVert_{2}^{2}),
%&= \frac{3}{2}\sigma_0^{2}\mathbf{v}_{ij}^{\mathrm{d}},
\end{aligned}
\end{equation}
% \end{align}
\noindent where (i) is from Theorem \ref{theo_main} since $\mathbf{Z}_X$ and $\mathbf{Z}_Y$ are independent. (ii) is from \eqref{Eq_ZxZyDefine}. (iii) $\mathbf{U}_{i,.}^{\mathrm{d}}$ and $\mathbf{V}_{j,.}^{\mathrm{d}}$ are the $i$ and $j$th row of $\mathbf{U}^{\mathrm{d}}$ and $\mathbf{V}^{\mathrm{d}}$. For conciseness, we define $\mathbf{v}_{ij}^{\mathrm{d}}=(\lVert\mathbf{U}_{i,.}^{\mathrm{d}}\lVert_{2}^{2}+\lVert\mathbf{V}_{j,.}^{\mathrm{d}}\lVert_{2}^{2})$. Similarly, the covariance between the element $(i,j)$ and $(m,n)$ is 

\begin{equation}
\begin{aligned}
% \begin{aligned}
&{\operatorname{Cov}}(\mathbf{S}_{i j}^{\sharp\mathrm{d}}-\hat{\mathbf{S}}_{i j}^{\sharp},\mathbf{S}_{m n}^{\sharp\mathrm{d}}-\hat{\mathbf{S}}_{m n}^{\sharp})\\
&=\frac{3}{2}\sigma_0^{2}[{\mathbf{e}_{j}}^{\transpose}(\mathbf{Y}^{\mathrm{d}} \mathbf{H}^{\mathrm{d}})(\mathbf{\Sigma}^{\mathrm{d}})^{-1}{(\mathbf{Y}^{\mathrm{d}} \mathbf{H}^{\mathrm{d}})^\transpose}{\mathbf{e}_{m}}\\
&+\mathbf{e}_{i}^{\transpose} (\mathbf{X}^{\mathrm{d}} \mathbf{H}^{\mathrm{d}})(\mathbf{\Sigma}^{\mathrm{d}})^{-1} (\mathbf{X}^{\mathrm{d}} \mathbf{H}^{\mathrm{d}})^\transpose\mathbf{e}_{n}].
%&{=} \frac{3}{2}\sigma_0^{2}(\mathbf{V}_{j,.}\mathbf{V}_{m,.}^\transpose+\mathbf{U}_{i,.}\mathbf{U}_{n,.}^\transpose).
% \end{aligned}
\end{aligned}
\end{equation}

It should be emphasized that there are two approximations in the process. One is ignoring the higher-order error term. Neglecting the higher-order error is regarded as the routine process in robotics \cite{barfoot2017state}, SLAM \cite{barfoot2014associating} and SfM \cite{andrew2001multiple}. The other approximation is ignoring the non-Gaussian noise $\mathbf{\Psi}_\mathrm{X}$ and $\mathbf{\Psi}_\mathrm{Y}$. We prove that the non-Gaussian residuals $\mathbf{\Psi}_\mathrm{X}$ and $\mathbf{\Psi}_\mathrm{Y}$ are always smaller than $\mathbf{Z}_X$ and $\mathbf{Z}_Y$. They do not have a heavy impact on the one-time Gaussian uncertainty propagation. The experiments also validate that the noise is predominated by Gaussian noises, and consequently, the non-Gaussian noise is negligible.\par

\subsection{Uncertainty for approximate low-rank structure}
%\subsection{Uncertainty propagation with the approximate low-rank structure}
\label{subsection_optimal_rank}

In the presence of noisy observations, an exact subspace projection can be used to mitigate the impact of noises. The original approximate SDP-based formulation \cite{dai2014simple} is mainly built for modelling noise-free observations. In solving \eqref{Eq_NRSfM_objective} contaminated with Gaussian noise, the original method \cite{dai2014simple} falls into the trap of over-fitting the noises. To efficiently handling this issue, the converged shape from the approximate SDP solver can be projected onto an exact low-rank subspace addressing the relation of shape and residual. With an optimal exact low-rank structure, the reprojection error should be consistent with the Gaussian noise.\par

The optimal rank selection is straightforward (Algorithm \ref{Algorithm_SDP}) and can be coupled with SDP-based NRSfM \cite{dai2014simple} as the post-processing module. The noise of the observation follows Gaussian distribution \eqref{Eq_observationGaussian}. We enforce this constraint by iteratively searching the optimal rank of $\mathbf{S}^{\mathrm{d}}$ whose reprojection residual $\lVert\mathbf{W}-\mathbf{R}\mathbf{S}^{\mathrm{d}}\lVert$ is closest to the prior distribution $\mathbf{W}$. Specifically, the testing rank $r$ is traversed from $1$ to maximum in searching for the optimal rank.\par

\subsection{Potential applications}

Uncertainty is indispensable in fusing multiple data sources. In contrast to batch optimization for dense or long trajectory cases, we demonstrate a preliminary test to segment the entire trajectory into several overlapping sub-trajectories. Each sub-trajectory is fed into SDP solver \cite{dai2014simple} individually (and in parallel with multi-core CPU). Then, all sub-trajectories are fused based on the estimated uncertainty. In this segmented process, the heavy time consumption on iterative SVD decomposition can be significantly reduced because the computation is on a much smaller matrix. Take SVD solver from \cite{cline2006computation} as an example, the time complexity is $\mathcal{O}\left(\min \left(m n^{2}, m^{2} n\right)\right)$ ($m$ and $n$ are matrix size). Therefore, the uncertainty estimation in this research enables the uncertainty-aware fusion of the overlapping sub-trajectories. \par

%增加我
%Algorithm \ref{Algorithm_SDP} presents the noise aware technical details for recovering the time-varying shape $\mathbf{S}^\mathrm{d}$ and the associated uncertainty from the observed 2D trajectories $\mathbf{W}$ based on the SDP formulation defined in \eqref{Eq_NRSfM_objective}. The minimizing process strictly follows the original SDP provided by \cite{dai2014simple}, while we couple it with the proposed noise-aware subspace process and the closed-form uncertainty propagation approach. Besides, following the works \cite{akhter2009nonrigid,agudo2017dust}, the camera projection matrix $\mathbf{R}$ is recovered independently. It is encoded as a standard least-square formulation and solved with the Gauss-Newton minimization, with the projection constraint $\mathbf{W}=\mathbf{R}{\mathbf{S}}$ and orthonormality constraint $\mathbf{I}_{2F}=\mathbf{R}\mathbf{R}^\transpose$. $\mathbf{I}_{2F}$ is the identity matrix with size $2F$.\par

\begin{table*}[]
	\caption{Demonstrated is the accuracy (average error) comparison between the original solver and the noise-aware solver. The recovered time-varying 3D shape is compared with the Monte Carlo's average.} % title of Table
	\centering
%	\begin{tabular}{c|c|c|c|c|c|c|c|c|c|c}
\setlength\tabcolsep{5.1pt} % default value: 6pt
	\begin{tabular}{c|cc|cc|cc|cc|cc|cc|cc}
\toprule
           & \multicolumn{2}{c|}{Drink}                                                             & \multicolumn{2}{c|}{Pick up}                                                           & \multicolumn{2}{c|}{Stretch}                                                           & \multicolumn{2}{c|}{Yoga}                                                              & \multicolumn{2}{c|}{Dance}                                                             & \multicolumn{2}{c|}{Paper}                                                             & \multicolumn{2}{c}{T-shirt}                                                            \\ \midrule \midrule
$\sigma_0$ & \multicolumn{1}{c|}{Original} & \begin{tabular}[c]{@{}c@{}}Noise-\\ aware\end{tabular} & \multicolumn{1}{c|}{Original} & \begin{tabular}[c]{@{}c@{}}Noise-\\ aware\end{tabular} & \multicolumn{1}{c|}{Original} & \begin{tabular}[c]{@{}c@{}}Noise-\\ aware\end{tabular} & \multicolumn{1}{c|}{Original} & \begin{tabular}[c]{@{}c@{}}Noise-\\ aware\end{tabular} & \multicolumn{1}{c|}{Original} & \begin{tabular}[c]{@{}c@{}}Noise-\\ aware\end{tabular} & \multicolumn{1}{c|}{Original} & \begin{tabular}[c]{@{}c@{}}Noise-\\ aware\end{tabular} & \multicolumn{1}{c|}{Original} & \begin{tabular}[c]{@{}c@{}}Noise-\\ aware\end{tabular} \\ \midrule \midrule
0.01       & \multicolumn{1}{c|}{0.0548}   & 0.0339                                                 & \multicolumn{1}{c|}{0.0751}   & 0.0601                                                 & \multicolumn{1}{c|}{0.0720}   & 0.0552                                                 & \multicolumn{1}{c|}{0.0452}   & 0.0622                                                 & \multicolumn{1}{c|}{0.1950}   & 0.1843                                                 & \multicolumn{1}{c|}{0.0749}   & 0.0572                                                 & \multicolumn{1}{c|}{0.0811}   & 0.0613                                                 \\
0.05       & \multicolumn{1}{c|}{0.1817}   & 0.0603                                                 & \multicolumn{1}{c|}{0.2374}   & 0.1130                                                 & \multicolumn{1}{c|}{0.2472}   & 0.1071                                                 & \multicolumn{1}{c|}{0.2216}   & 0.1016                                                 & \multicolumn{1}{c|}{0.2162}   & 0.3352                                                 & \multicolumn{1}{c|}{0.1218}   & 0.0832                                                 & \multicolumn{1}{c|}{0.1944}   & 0.0694                                                 \\
0.08       & \multicolumn{1}{c|}{0.2821}   & 0.0785                                                 & \multicolumn{1}{c|}{0.3372}   & 0.1617                                                 & \multicolumn{1}{c|}{0.3404}   & 0.1364                                                 & \multicolumn{1}{c|}{0.3005}   & 0.1284                                                 & \multicolumn{1}{c|}{0.4419}   & 0.2482                                                 & \multicolumn{1}{c|}{0.2264}   & 0.0969                                                 & \multicolumn{1}{c|}{0.2538}   & 0.0827                                                 \\
0.10       & \multicolumn{1}{c|}{0.3502}   & 0.0807                                                 & \multicolumn{1}{c|}{0.4525}   & 0.1587                                                 & \multicolumn{1}{c|}{0.4756}   & 0.1632                                                 & \multicolumn{1}{c|}{0.4208}   & 0.1504                                                 & \multicolumn{1}{c|}{0.5194}   & 0.2615                                                 & \multicolumn{1}{c|}{0.2588}   & 0.1115                                                 & \multicolumn{1}{c|}{0.3011}   & 0.1001                                                 \\
0.20       & \multicolumn{1}{c|}{0.5615}   & 0.1218                                                 & \multicolumn{1}{c|}{0.8886}   & 0.2605                                                 & \multicolumn{1}{c|}{0.8096}   & 0.2329                                                 & \multicolumn{1}{c|}{0.7216}   & 0.2303                                                 & \multicolumn{1}{c|}{0.9501}   & 0.3560                                                 & \multicolumn{1}{c|}{0.3687}   & 0.1361                                                 & \multicolumn{1}{c|}{0.4399}   & 0.1301                                                 \\ 
\bottomrule
\end{tabular}
	\label{table_accuracy}
\end{table*}

\begin{table*}[]
	\caption{The coverage rates of $1.96\sigma_0
    \sqrt{\frac{3}{2}\mathbf{v}_{ij}^{\mathrm{d}(k)}}$ bound in each trial of the 100 Monte Carlo. It has been repeated for different $\sigma_0$. Since the element-wise coverage rate cannot be presented, we count the mean and standard deviation (`std') for each Monte-Carlo test. The proposed closed-form uncertainty propagation approach describes the probability distribution well. The noise is proportional to the shape because the shape has already been normalized to $[0 -  1]$.} % title of Table
	\centering
%	\begin{tabular}{c|c|c|l|l|l|l|l|l|l|l}
\setlength\tabcolsep{2.7pt} % default value: 6pt
	\begin{tabular}{c|cc|cc|cc|cc|cc|cc|cc|cc|cc}
\toprule
           & \multicolumn{2}{c|}{Drink}           & \multicolumn{2}{c|}{Pick up}         & \multicolumn{2}{c|}{Stretch}         & \multicolumn{2}{c|}{Yoga}            & \multicolumn{2}{c|}{Dance}           & \multicolumn{2}{c|}{Paper}           & \multicolumn{2}{c|}{T-shirt}         & \multicolumn{2}{c|}{Face3}           & \multicolumn{2}{c}{Face4}            \\ \midrule \midrule
$\sigma_0$ & \multicolumn{1}{c|}{Mean}   & Std    & \multicolumn{1}{c|}{Mean}   & Std    & \multicolumn{1}{c|}{Mean}   & Std    & \multicolumn{1}{c|}{Mean}   & Std    & \multicolumn{1}{c|}{Mean}   & Std    & \multicolumn{1}{c|}{Mean}   & Std    & \multicolumn{1}{c|}{Mean}   & Std    & \multicolumn{1}{c|}{Mean}   & Std    & \multicolumn{1}{c|}{Mean}   & Std    \\ \midrule \midrule
0.01       & \multicolumn{1}{c|}{0.9632} & 0.0268 & \multicolumn{1}{c|}{0.9343} & 0.0489 & \multicolumn{1}{c|}{0.9324} & 0.0359 & \multicolumn{1}{c|}{0.9166} & 0.0612 & \multicolumn{1}{c|}{0.9249} & 0.0498 & \multicolumn{1}{c|}{0.9413} & 0.0272 & \multicolumn{1}{c|}{0.9398} & 0.0340 & \multicolumn{1}{c|}{0.9342} & 0.0381 & \multicolumn{1}{c|}{0.9281} & 0.0392 \\
0.05       & \multicolumn{1}{c|}{0.9661} & 0.0272 & \multicolumn{1}{c|}{0.9382} & 0.0402 & \multicolumn{1}{c|}{0.9394} & 0.0398 & \multicolumn{1}{c|}{0.9308} & 0.0437 & \multicolumn{1}{c|}{0.9375} & 0.0420 & \multicolumn{1}{c|}{0.9427} & 0.0266 & \multicolumn{1}{c|}{0.9428} & 0.0311 & \multicolumn{1}{c|}{0.9366} & 0.0360 & \multicolumn{1}{c|}{0.9369} & 0.0386 \\
0.10       & \multicolumn{1}{c|}{0.9623} & 0.0306 & \multicolumn{1}{c|}{0.9431} & 0.0365 & \multicolumn{1}{c|}{0.9441} & 0.0388 & \multicolumn{1}{c|}{0.9398} & 0.0416 & \multicolumn{1}{c|}{0.9416} & 0.0364 & \multicolumn{1}{c|}{0.9455} & 0.0210 & \multicolumn{1}{c|}{0.9436} & 0.0285 & \multicolumn{1}{c|}{0.9396} & 0.0302 & \multicolumn{1}{c|}{0.9388} & 0.0327 \\
0.20       & \multicolumn{1}{c|}{0.9667} & 0.0221 & \multicolumn{1}{c|}{0.9535} & 0.0296 & \multicolumn{1}{c|}{0.9477} & 0.0415 & \multicolumn{1}{c|}{0.9400} & 0.0415 & \multicolumn{1}{c|}{0.9463} & 0.0329 & \multicolumn{1}{c|}{0.9471} & 0.0196 & \multicolumn{1}{c|}{0.9433} & 0.0254 & \multicolumn{1}{c|}{0.9408} & 0.0295 & \multicolumn{1}{c|}{0.9403} & 0.0345 \\ \bottomrule
\end{tabular}
	\label{table_comparison_of_coverage_ratio}
\end{table*}

\begin{table}[]
	\caption{The probability values ($p$-value) of the Shapiro-Wilk test of $\mathbf{S}^{\mathrm{d}}_{1}$, $\mathbf{S}^{\mathrm{d}}_{2}$, $\mathbf{S}^{\mathrm{d}}_{3}$, $\mathbf{S}^{\mathrm{d}}_{4}$ and $\mathbf{S}^{\mathrm{d}}_{5}$. The $p$-value $> 0.05$ indicates a significant possibility that the element is normally distributed.} % title of Table
	\centering
	\begin{tabular}{p{0.57cm}<{\raggedleft}|p{1.11cm}<{\centering}|p{1.11cm}<{\centering}|p{1.11cm}<{\centering}|p{1.11cm}<{\centering}|p{1.11cm}<{\centering}}
		\toprule 
		% \diagbox[width=6.5em,trim=l]{$\delta_0$}{Element} & $\mathbf{S}^{\mathrm{d}}_{1}$ & $\mathbf{S}^{\mathrm{d}}_{2}$ & $\mathbf{S}^{\mathrm{d}}_{3}$ & $\mathbf{S}^{\mathrm{d}}_{4}$ & $\mathbf{S}_{2560}$ \\ \midrule \midrule
		$\delta_0$ & $\mathbf{S}^{\mathrm{d}}_{1}$ & $\mathbf{S}^{\mathrm{d}}_{2}$ & $\mathbf{S}^{\mathrm{d}}_{3}$ & $\mathbf{S}^{\mathrm{d}}_{4}$ & $\mathbf{S}^{\mathrm{d}}_{5}$ \\ \midrule \midrule
		0.01                                                                     & 0.1962           & 0.1258            & 0.6253            & 0.0361            & 0.7208              \\
		0.05                                                                     & 0.0766            & 0.2938            & 0.7607            & 0.3681            & 0.1720              \\
		0.08                                                                     & 0.0538            & 0.5320            & 0.0836            & 0.9540            & 0.9793              \\
		0.10                                                                      & 0.0960            & 0.9755            & 0.4643            & 0.0090            & 0.1383              \\ \bottomrule
	\end{tabular}
	\label{table_wilk_shapiro_test}
\end{table}
\begin{table}[bthp]
	\caption{Presented is the robustness test of the coverage rate over different rank. We exert $\mathrm{m}\%, \mathrm{m}\in [-10,-20,10,20]$ percentage error over the estimated rank and test the coverage rates of $1.96\sigma_0
    \sqrt{\frac{3}{2}\mathbf{v}_{ij}^{\mathrm{d}(k)}}$ bound for different $\sigma_0$ over 100 Monte Carlo trials. } % title of Table
	\centering
%	\begin{tabular}{c|c|c|c|c|c|c|c|c}
	\begin{tabular}{p{0.36cm}<{\raggedleft}|p{0.56cm}<{\centering}|p{0.56cm}<{\centering}|p{0.56cm}<{\centering}|p{0.56cm}<{\centering}|p{0.56cm}<{\centering}|p{0.56cm}<{\centering}|p{0.56cm}<{\centering}|p{0.56cm}<{\centering}}
		\toprule 
		& \multicolumn{2}{c|}{+10\%} & \multicolumn{2}{c|}{+20\%} & \multicolumn{2}{c|}{-10\%} & \multicolumn{2}{c}{-20\%} \\ \midrule \midrule
		$\sigma_0$ & Mean         & Std         & Mean         & Std         & Mean         & Std         & Mean         & Std         \\ \midrule \midrule
		0.01       & 0.932       & 0.055      & 0.937       & 0.054      & 0.911       & 0.065      & 0.886       & 0.079      \\
		0.05       & 0.942       & 0.039      & 0.948       & 0.038      & 0.938       & 0.042      & 0.895       & 0.070      \\
		0.08       & 0.948       & 0.035      & 0.948       & 0.035      & 0.941       & 0.038      & 0.878       & 0.085      \\
		0.10       & 0.947       & 0.037      & 0.948       & 0.036      & 0.941       & 0.040      & 0.875       & 0.091      \\
		0.20       & 0.950       & 0.034      & 0.952       & 0.033      & 0.949       & 0.034      & 0.897       & 0.075      \\ \bottomrule
	\end{tabular}
	\label{table_robustness_analysis}
\end{table}

\begin{algorithm}[t]
	\caption{Rank searching for noisy observations.}
	\label{Algorithm_SDP}
	%\LinesNumbered %
	\KwIn{${\mathbf{W}}$, $\sigma_0$, $\mathbf{S}^{\sharp\mathrm{d}}$ from SDP}
	\KwOut{Optimal rank $r$ and exact version of $\mathbf{S}^{\sharp\mathrm{d}}$}
	$r=1$\\
	%$r=\operatorname{Rank}(\mathbf{S}^{\sharp\mathrm{d}})$\\
	\nextnr
	\While{$r<\operatorname{max}(3N,F)$}{
		\nextnr
		$\mathbf{S}^{\sharp\mathrm{d}}=g(\pi_r (g^{-1}(\mathbf{S}^{\sharp\mathrm{d}})))$\\ 
		\nextnr
		$r=r+1$\\
		\nextnr
		/* Check Convergence */  \\
		\nextnr                
		$95\%$ elements in $\!\lVert\mathbf{W}\!-\!\mathbf{R}g^{-1}\!(\mathbf{S}^{\sharp\mathrm{d}})\!\lVert\!$ within $1.96\sigma_0$\\
	}
	Notation: $\pi_r(\cdot)$ projects to exact rank with hinge loss.\\
\end{algorithm}

\section{Results and discussion}
\label{Section_results}

Similar to previous works \cite{dai2014simple,cabral2013unifying,agudo2017dust}, the proposed method has been validated quantitatively on the classic MoCap data set \cite{MoCap} which uses 41 markers distributed on the surface of the human body. Five sequences were chosen and their 3D points were projected onto the 2D space with a virtual orthographic camera following a relative circular motion around the object at a stable angular speed. The five data sets were: `Drink', `Pickup', `Yoga', `Stretch' and `Dance'. Two real-world dense deforming `paper' and `T-shirt' datasets obtained by Kinect with 3D ground truth were tested \cite{varol2012constrained}. Validity masks were enforced on them. Moreover, two dense face data sets from \cite{sidhu2020neural} with ground truth were also adopted to test the performance. The ground truth rotations were directly adopted for the dense data sets. All shapes were normalized to the range $[0\ 1]$. All computations were conducted on a commercial desktop with CPU i5-9400 and Matlab 2020a.\par %Fig. \ref{fig_mocapresult} shows some sample results of the reconstructions.\par

%下面的完整版
%The proposed method was validated quantitatively on several scenarios in the MoCap data set provided by CMU \cite{MoCap}. The MoCap data set uses more than 30 markers distributed on the skeleton of the human body. An infrared camera tracks accurate 3D marker positions of human body motions. It is a widely used data set for validation in computer vision community \cite{agudo2017dust,dai2014simple,cabral2013unifying}. Five widely tested data sets used in \cite{akhter2009defense} were adopted, which project the original 3D points onto the 2D space with an orthographic camera following a relative circular motion around the object, at a stable angular speed. The five data sets include: `Drink' (1102/41), `Pickup (357/41)', `Yoga (307/41)', `Stretch (370/41)' and `Dance (528/75)'. Where ($F$/$N$) is the number of frames $F$ and the number of tracked points $N$. For the benefit of quantifying the level of noise, all shapes are normalized to the range $[0\ 1]$.\par %Fig. \ref{fig_mocapresult} shows some sample results of the reconstructions.\par

Table \ref{table_accuracy} reveals that the proposed exact rank searching makes the SDP-based formulation more robust to noisy observations as claimed in Section \ref{subsection_optimal_rank}. It indicates that the proposed noise-aware exact rank searching (Algorithm \ref{Algorithm_SDP}) is necessary for noisy observations (at least more than $1\%$ standard deviation). We project the approximate $\mathbf{S}^{\sharp\mathrm{d}}$ to the subspace correctly. The proposed noise-aware module helps the SDP-based NRSfM achieves better accuracy in the presence of the Gaussian noises. In experiment, we also notice that projecting the time-varying shape onto subspace with a smaller rank achieves better accuracy, especially with heavy noises.
%Following earlier works \cite{akhter2009nonrigid,dai2014simple,agudo2017dust}, the overall error is measured as:\par

%用我替换 过去时
%Results reveal that the proposed method is more robust to noisy observation. Table \ref{table_accuracy} indicates that the proposed noise-aware algorithm (Algorithm \ref{Algorithm_SDP}) achieves better accuracy in the presence of the noise. Projecting the model into subspace with a smaller rank achieves better accuracy, especially in heavy noise scenario.\par

% \begin{equation}
% \label{Eq_error_index}
% \mathrm{e}_s=\frac{1}{\delta F N}\sum_{i=1}^F\sum_{j=1}^N\mathrm{e}^i_{j},\  \delta=\frac{1}{3F}\sum_{i=1}^F(\delta_x^i+\delta_y^i+\delta_z^i),\
% \end{equation}
% \noindent where $\mathrm{e}^i_{j}$ is the Euclidean distance error of the point and ${\delta}_x^i$, ${\delta}_y^i$ and ${\delta}_z^i$ are the standard deviations of each point at frame $i$.\par 

\subsection{Aleatoric uncertainty propagation}
Monte Carlo tests were conducted on the 2D observations with different levels of noise. For each test, Gaussian noise was imposed on the 2D tracked points with different standard deviations $\sigma_0$. All parameter settings STRICTLY followed the work \cite{dai2014simple}. We also tested the augmented Lagrange multiplier solver used in \cite{agudo2017dust} (our implementation). All the numerical results from Lagrange multiplier and augmented Lagrange multiplier solvers are very close since the cost functions are the same in \eqref{Eq_NRSfM_objective}. Thus, only the results from the Lagrange multiplier \cite{dai2014simple} are presented. Besides, we need to clarify that most state-of-the-art SDP-based algorithms are not open-sourced and it is difficult to test more.
%including ${i}_{max}=20$, $\mu=4$, $\eta_{\mu}=0.25$ and $\epsilon=10^{-10}$.
Monte Carlo test of $\mathrm{T}=100$ times were performed on each data set with the given $\sigma_0$. Since the noise-free estimation $\hat{\mathbf{S}}^{\sharp}$ from Algorithm \ref{Algorithm_SDP} suffers from local minima, the Monte Carlo's average $\overline{\mathbf{S}}^{\sharp\mathrm{d}}$ is used instead. The element-wise error in trial $k$ is defined as

\begin{equation}
\mathrm{e}^{(k)}_{i j}={\mathbf{S}_{i j}^{\sharp\mathrm{d}(k)}}-\overline{\mathbf{S}}_{i j}^{\sharp\mathrm{d}(k)},\  k\in [1,\cdots,100].
\end{equation} 

We intend to validate that the estimated variance $\frac{3}{2}\sigma_0^{2}\mathbf{v}_{ij}^{\mathrm{d}}$ is correct, that is, $95\%$ of $\mathrm{e}^{(k)}_{i j}$ fall within the range of $\left[ - 1.96\sigma_0
\sqrt{\frac{3}{2}\mathbf{v}_{ij}^{\mathrm{d}(k)}},\ 1.96\sigma_0
\sqrt{\frac{3}{2}\mathbf{v}_{ij}^{\mathrm{d}(k)}}\right]$ ($\mathbf{v}_{ij}^{\mathrm{d}(k)}$ is the $k$th trial of $\mathbf{v}_{ij}^{\mathrm{d}}$). Thus, define the \textbf{coverage rate} as the ratio of errors fall in the $1.96$ bound. Table \ref{table_comparison_of_coverage_ratio} shows the general statistical results over all elements of the 5 data sets, and it indicates that the closed-form uncertainty quantification coverage rate is close to $95\%$. The mean and standard deviation count the element-wise coverage rate since the element-wise coverage rates cannot be presented individually \footnote{Readers are encouraged to test the provided code.} Table \ref{table_comparison_of_coverage_ratio} shows that the coverage rate is close to $95\%$ and is more accurate in large $\sigma_0$. Thus, the impact of the non-Gaussian residuals $\mathbf{\Psi}_\mathrm{X}$ and $\mathbf{\Psi}_\mathrm{Y}$ is small regarding the Gaussian residuals. Fig. \ref{fig_error_ellipse} and the attached video are provided for better visualization of the error ellipisoid with $1.96\sigma_0$ bound ($\sigma_0=0.05$). Finally, it should be addressed that the uncertainty is aleatoric and should be validated with Monte Carlo test (routine in SLAM and SfM). Comparing directly with ground truth involves epistemic and structure uncertainty. \par 

\begin{figure}[t]
	\centering
	\subfloat{
		\begin{minipage}[]{0.5\textwidth}
			\centering
			\includegraphics[width=1\linewidth]{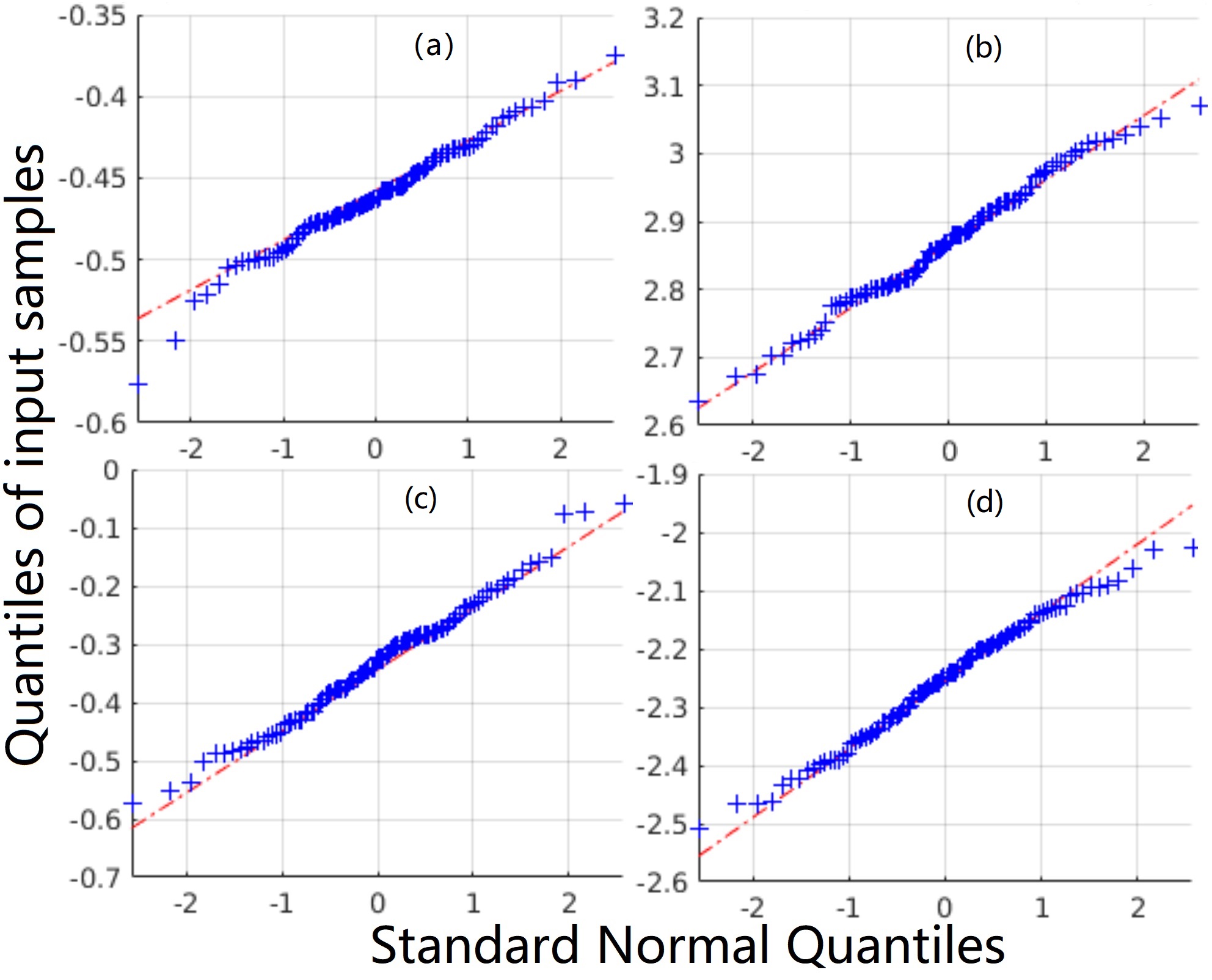}
		\end{minipage}
	}
	%\caption{The Q-Q plots of $\mathbf{S}^{\mathrm{d}}_{1}$, $\mathbf{S}^{\mathrm{d}}_{2}$, $\mathbf{S}^{\mathrm{d}}_{3}$ and $\mathbf{S}^{\mathrm{d}}_{4}$. 100 independent trials are conducted for noise level $\sigma_0=0.05$. }
	\caption{Illustrated are the Q-Q plots of $\mathbf{S}^{\mathrm{d}}_{1}$, $\mathbf{S}^{\mathrm{d}}_{2}$, $\mathbf{S}^{\mathrm{d}}_{3}$ and $\mathbf{S}^{\mathrm{d}}_{4}$ against the standard normal distribution in (a), (b), (c) and (d) respectively. They are over 100 independent trials for noise $\sigma_0=0.05$ in the four data sets. }
	\label{fig_3data sets}
\end{figure}

In addition to the general coverage rate of the proposed aleatoric uncertainty propagation approach, we also validate that the error of the estimated shape ${\mathbf{S}^{\sharp\mathrm{d}}}$ follows the Gaussian distribution. The Quantile-Quantile (Q-Q) plots (Fig. \ref{fig_3data sets}) of $\mathbf{S}^{\mathrm{d}}_{1}$ (element (1,1) of `drink'), $\mathbf{S}^{\mathrm{d}}_{2}$ (element (3,4) of `pickup'), $\mathbf{S}^{\mathrm{d}}_{3}$ (element (5,8) of `stretch'), $\mathbf{S}^{\mathrm{d}}_{4}$ (element (30,30) of `yoga') and $\mathbf{S}^{\mathrm{d}}_{5}$ (element (25,60) `yoga') are selected randomly and presented in the case of $\sigma_0=0.05$ against the standard normal distribution. Table \ref{table_wilk_shapiro_test} is presented to show that the errors of the recovered 3D points follow the normal distribution. The $p$-value reveals that 15 out of 16 samples are significant in the normal distribution ($> 0.05$). Since all elements in the recovered shape should be consistent (either obey or reject the normal distribution), we can conclude that the shape noise is predominated by Gaussian noise and the non-Gaussian residuals $\mathbf{\Psi}_\mathrm{X}$ and $\mathbf{\Psi}_\mathrm{Y}$ are trivial. \par
%The only samples that reject this assumption in higher possibility may be attributed to a small sample of training sets.\par  

% \begin{figure}[!h]
% 	\centering
% 	\subfloat{
% 		\begin{minipage}[]{0.13\textwidth}
% 			\centering
% 			\includegraphics[width=1\linewidth]{qqplot_1x}
% 		\end{minipage}
% 	}
% 	\subfloat{
% 		\begin{minipage}[]{0.13\textwidth}
% 			\centering
% 			\includegraphics[width=1\linewidth]{qqplot_2x}
% 		\end{minipage}
% 	}\/
% 	\subfloat{
% 		\begin{minipage}[]{0.13\textwidth}
% 			\centering
% 			\includegraphics[width=1\linewidth]{qqplot_3x}
% 		\end{minipage}
% 	}
% 	\subfloat{
% 		\begin{minipage}[]{0.13\textwidth}
% 			\centering
% 			\includegraphics[width=1\linewidth]{qqplot_4x}
% 		\end{minipage}
% 	}
% 	\caption{Illustrated are the Q-Q (quantile-quantile) plots of $\mathbf{S}^{\mathrm{d}}_{1}$, $\mathbf{S}^{\mathrm{d}}_{2}$, $\mathbf{S}^{\mathrm{d}}_{3}$ and $\mathbf{S}^{\mathrm{d}}_{4}$ against the standard normal distribution in (a), (b)
% 		,(c) and (d) respectively. They are over 100 independent trials for noise $\sigma_0=0.05$ in the four data sets. }
% 	\label{fig_3data sets}
% \end{figure}

\subsection{Robustness of the approximate structure}
Apart from the coverage rate tests, we further validate that the proposed approach is robust to the numerical rank selection process. Numerically, $\mathbf{v}_{ij}^\mathrm{d} \in [0,2]$ is dependent on the rank selection, as it is $0$ when the matrix is null and $2$ when it is full-rank. Different ranks in Algorithm \ref{Algorithm_SDP} were tested and presented in table \ref{table_robustness_analysis}.\par

Table \ref{table_robustness_analysis} illustrates the coverage rate with different disturbances on the rank numerically estimated in Algorithm \ref{Algorithm_SDP}. Among the 16 tests, 9 achieves $2\%$ error, and 12 achieves $3\%$ errors regarding Monte Carlo average's $95\%$ coverage rate. Moreover, the standard deviations do not increase significantly with larger rank disturbance in Table \ref{table_robustness_analysis}. Therefore, the presented uncertainty propagation approach is robust to the optimal rank selection in the numerical estimation process. The standard deviations reveal the robustness to subspace rank selection.\par

% \begin{table}[]
% 	\caption{Presented are the ratios between the average variance of the small parallax point and the average of the rest normal points.} % title of Table
% 	\centering
% 	\begin{tabular}{p{0.57cm}<{\raggedleft}|p{1.11cm}<{\centering}|p{1.11cm}<{\centering}|p{1.11cm}<{\centering}|p{1.11cm}<{\centering}|p{1.11cm}<{\centering}}
% 		\toprule 
% 		& Drink  & Pick up & Stretch & Yoga   & Dance  \\ \midrule \midrule
% 		Ratio & 7.3402 & 4.2134  & 4.3921  & 9.7167 & 3.0740 \\ \bottomrule
% 	\end{tabular}
% 	\label{table_smallparallax}
% \end{table}

\begin{table}[]
\caption{The comparisons between the original work \cite{dai2014simple} and the overlapping sub-trajectory fusion. The optimized sub-trajectories are fused based on the closed-form uncertainty. Time is measured in $1000$s. Time$^{*}$ is obtained by parallel computing on 6 cores CPU.}
%\begin{tabular}{c|c|c|c|c|c}
\begin{tabular}{p{0.84cm}<{\centering}|p{1.3cm}<{\centering}|p{0.94cm}<{\centering}|p{1.3cm}<{\centering}|p{0.94cm}<{\centering}|p{0.94cm}<{\centering}}
\toprule
               & \multicolumn{2}{c|}{Original} & \multicolumn{3}{c}{Uncertainty-aware} \\ \midrule \midrule
               & Accuracy        & Time        & Accuracy                      & Time                       & Time$^{*}$                       \\ \midrule \midrule
Face3 &     0.052     &   7.363   & 0.053                        & 5.978                      & 1.533                      \\
Face4 &   0.044        &   6.972   &      0.044             &  5.193                  &    1.266             \\
\bottomrule
% Drink          &    0.0225     &             &    0.0218                  &       0.472              &          0.0911                  \\ 
%Pick up          &    0.0486      &             &  0.0435                    &         0.112           &       0.024             \\\hline
\end{tabular}
\label{table:sub_fusion}
\end{table}

\subsection{Potential applications}  

One potential application of the confidence quantification is data fusion. To reduce the time consumption, the entire trajectory can be divided into several submaps and joined after processing \cite{wang2019submap}. Two dense face data sets from \cite{sidhu2020neural} with ground truth were adopted. Each sequential data set was divided into 6 sub-trajectories with $20\%$ overlapping and processed in parallel with 6 CPU cores. Table \ref{table:sub_fusion} shows that parallel processing is much faster than batch optimization. The accuracy of the sub-trajectories fusion is similar to SDP in a batch. Further tests indicate that the accuracy of the uncertainty-aware fused shape on the overlapping regions is from $3\%$ to $5\%$ higher than the average fusion. \par

% The uncertainty quantification also helps identifying the degenerated features with small parallax. When the observations share same uncertainty, small parallax affects the accuracy of the prediction. To analyze this issue, one feature point with small parallax was synthesized to the MoCap data sets, and the ratio of the uncertainty was tested on other 3D points. Note that the variances of the observation are all fixed to $\sigma_0$. Table \ref{table_smallparallax} shows that small parallax features have larger variances.\par 

\section{Conclusion and future works}
%用我替换
% We propose a closed-form element-wise uncertainty propagation algorithm for the state-of-the-art low-rank based NRSfM problem. This is the first research addressing the uncertainty quantification issue in this field. The proposed method only requires the statistical distribution of the 2D observation. To overcome the approximate low-rank structure of the time-varying shape, we present a numerical approach to estimate the optimal rank and convert the problem from approximate low-rank recovery to exact low-rank recovery. Extensive experiments show that our closed-form uncertainty propagation approach accurately describes the results of SDP-based NRSfM.\par  
This is the first research to address the uncertainty propagation in SDP-based NRSfM problems. We propose a closed-form element-wise uncertainty propagation algorithm for the state-of-the-art SDP-based approach. The proposed method only requires the statistical distribution of the errors in 2D observation. Monte Carlo tests validate that the coverage rate of the element-wise variance precisely describes the estimated time-varying shape distribution. Robustness tests demonstrate that the uncertainty is not sensitive to the rank estimated with our modified SDP workflow. Our closed-form uncertainty propagation approach may benefit applications like parallel processing of SDP-based methods.\par  

Although the uncertainty cannot directly improve the accuracy, it allows a more statistically sound way to utilize the estimation
results, such as parallel processing, fusion with other sensors or risk evaluation of the system in a safety-critical situation (e.g., autonomous vehicle). Future work may focus on exploiting solutions to estimate uncertainty quantification for 2D trackings with missing or occluded elements.\par

\section*{Acknowledgment}
Toyota Research Institute provided funds to support this work. Funding for M. Ghaffari was in part
provided by NSF Award No. 2118818.

%\appendix

\section*{Appendix. }
\label{Section_appendix}

% \begin{remark}
% \label{remark_infinite_orthogonal}
% Given nonzero full rank matrices $\mathbf{P} \in \mathbb{R}^{n \times r}$ and $\mathbf{B} \in \mathbb{R}^{n \times r}$, there exists infinite $n \times n$ orthogonal matrix $\mathbf{M}$ satisfying $\mathbf{M}\mathbf{P}=\mathbf{B}$ if $\frac{n(n-1)}{2} \geq r$. It is because the $n \times n$ orthogonal matrix has $\frac{n(n-1)}{2}$ DoF \cite{aitken2017determinants} and the number of $n$ dimensional point $\mathbf{P}$ is $r$.
% \end{remark}

The appendix is the proof of Theorem \ref{theo_main}. This section first explicitly derives the relation between $\mathbf{S}^\sharp$ and $\mathbf{R}$ in \eqref{Eq_NRSfM_objective}. Then, the uncertainties of $\mathbf{X}^\mathrm{d}$ and $\mathbf{Y}^\mathrm{d}$ are inferred taking advantage of the zero derivatives of \eqref{Eq_NRSfM_objective} at its local minima ($\mathbf{X}^\mathrm{d}$ and $\mathbf{Y}^\mathrm{d}$).

\begin{remark}
\label{remark_var_prop}
Given two i.i.d. variables $\mathrm{x} {\sim} \mathcal{N}(\overline{\mathrm{x}}, \delta_{\mathrm{x}}^{2})$ and $\mathrm{y} {\sim} \mathcal{N}(\overline{\mathrm{y}}, \delta_{\mathrm{y}}^{2})$, the product $\mathrm{x}\mathrm{y}$'s variance $\delta_{\mathrm{x}\mathrm{y}}^2=\overline{\mathrm{x}}^2\delta_{\mathrm{y}}^{2}+\overline{\mathrm{y}}^2\delta_{\mathrm{x}}^{2}$. If $\overline{\mathrm{y}}=0$, the variance degrades to $\delta_{\mathrm{x}\mathrm{y}}^2=\overline{\mathrm{x}}^2\delta_{\mathrm{y}}^{2}$. 
\end{remark}

\begin{remark}
\label{remark_upperbound}
Denote the non-Gaussian noise of $\mathbf{G}_1^\sharp=\mathbf{X}^{\mathrm{d}}\mathbf{H}{\hat{\mathbf{Y}}^{\transpose}}-{\hat{\mathbf{X}}}{\hat{\mathbf{Y}}^{\transpose}}$ and $\mathbf{G}_2^\sharp=\hat{\mathbf{X}}\mathbf{H}^\transpose{\mathbf{Y}^{\mathrm{d}\transpose}}-{\hat{\mathbf{X}}}{\hat{\mathbf{Y}}^{\transpose}}$ resulted from the local optimization. Since shape noise $\mathbf{E}^\sharp=\mathbf{X}^{\mathrm{d}}{\mathbf{Y}^{\mathrm{d}\transpose}}-{\hat{\mathbf{X}}}\hat{\mathbf{Y}}^{\transpose}$ and the global rectification \eqref{Eq_rectify}, the non-Gaussian noises $\mathbf{G}_1^\sharp$ and $\mathbf{G}_2^\sharp$ are always smaller than $\mathbf{E}^\sharp$.
\end{remark}

\begin{remark}
\label{remark_U_expect}
Given a random orthogonal matrix $\mathbf{U}_i \in \mathcal{A}^{3N \times 3N}$, the element-wise expected value $\mathcal{E}(\mathbf{U}_i)=\frac{1}{\sqrt{3N}}\mathbf{1}$, where $\mathbf{1}$ is the all 1 matrix with size $3N\times 3N$. For the matrix $\dot{\mathbf{U}}_i$ which is composed of the first $2N$ rows of $\mathbf{U}_i$, $\mathcal{E}(\dot{\mathbf{U}}_i^\transpose\dot{\mathbf{U}}_i)=\mathrm{p}\mathbf{I}_{3N}$ and $\mathrm{p}=\frac{2}{3}$.
\end{remark}

%Denote $\tilde{\mathbf{R}}_i$ as the full $3 \times 3$ rotation matrix of $\mathbf{R}_i$, 
We next analyze $g(\cdot)$ and convert ${\mathbf{R}}\mathbf{S}$ to an explicit form of $\mathbf{S}^\sharp$ projection. Denote $\tilde{\mathbf{R}}$ as the full $3F\times3F$ of $\mathbf{R}$, $\tilde{\mathbf{W}}=\tilde{\mathbf{R}}\mathbf{S}$ and $\tilde{\mathbf{S}}^\sharp = g(\tilde{\mathbf{R}}\mathbf{S})$. $\tilde{\mathbf{S}}^\sharp$ is the frame-wise rotated ${\mathbf{S}}^\sharp$. With some manipulation, we have

\begin{equation}
\tilde{\mathbf{S}}^\sharp=
\begin{bmatrix}
\Pi(\tilde{\mathbf{R}}_1)\hat{\mathbf{X}}\hat{\mathbf{Y}}_{1,.}^\transpose,	& \cdots, & \Pi(\tilde{\mathbf{R}}_F) \hat{\mathbf{X}}\hat{\mathbf{Y}}_{F,.}^\transpose
\end{bmatrix}
\approx\mathbf{U}_{\Delta}\hat{\mathbf{X}}\hat{\mathbf{Y}}^\transpose,	 
\end{equation}

\begin{equation}
\label{Eq_new_R}
\Pi(\tilde{\mathbf{R}}_i)=
\begin{bmatrix}
\mathbf{I}_N	\tilde{\mathbf{R}}_{i(11)}&\mathbf{I}_N	\tilde{\mathbf{R}}_{i(12)} & \mathbf{I}_N	\tilde{\mathbf{R}}_{i(13)}\\
\mathbf{I}_N	\tilde{\mathbf{R}}_{i(21)}&\mathbf{I}_N	\tilde{\mathbf{R}}_{i(22)} & \mathbf{I}_N	\tilde{\mathbf{R}}_{i(23)}\\
\mathbf{I}_N	\tilde{\mathbf{R}}_{i(31)}&\mathbf{I}_N	\tilde{\mathbf{R}}_{i(32)} & \mathbf{I}_N	\tilde{\mathbf{R}}_{i(33)}\\
\end{bmatrix},	 
\end{equation}

\noindent where $\mathbf{U}_{\Delta}\in 3N\times3N$ is the orthogonal matrix satisfying all $\Pi(\tilde{\mathbf{R}}_i), i \in [1,F]$. The orthogonal matrix $\Pi(\tilde{\mathbf{R}}_i) \in 3N\times3N$ is defined in \eqref{Eq_new_R}. $\tilde{\mathbf{R}}_{i(uv)}$ is the $(u,v)$ element of $\tilde{\mathbf{R}}_i$. $\mathbf{I}_N$ is the identity matrix with size $N$. Next, we discuss the existence of $\mathbf{U}_{\Delta}$.\par

\begin{remark}
\label{lemma_infiite_solu}
On the manifold of all factorization estimators of $\tilde{\mathbf{S}}^\sharp$, there are infinite exact $\mathbf{U}_{\Delta}$ on condition that $\frac{3N(3N-1)}{2} > F$, one exact $\mathbf{U}_{\Delta}$ if $\frac{3N(3N-1)}{2} = F$ and an optimial approximate $\mathbf{U}_{\Delta}$ if $\frac{3N(3N-1)}{2} < F$. For all three situations, with the sample $\mathbf{U}_{\Delta}$ large enough (or frame numbers $F$), the expectant  $\mathcal{E}(\dot{\mathbf{U}}_{\Delta}^\transpose\dot{\mathbf{U}}_{\Delta})=\mathrm{p}\mathbf{I}_{3N}$ ($\mathrm{p}=\frac{2}{3}$) and $\mathcal{E}(\dot{\mathbf{U}}_{\Delta})=\frac{1}{\sqrt{3N}}\mathbf{1}$. This follows remark \ref{remark_U_expect}. $\dot{\mathbf{U}}_{\Delta}$ is the first $2N$ rows of $\mathbf{U}_{\Delta}$. The accuracy is dependent on the number of frames $F$.
\end{remark}

% \begin{proof}
% Applying $\mathbf{X} := \mathbf{U} \mathbf{\Sigma}^{1 / 2}$, the lemma holds if there exists infinite  orthogonal matrix $\mathbf{U}' \in \mathcal{A}^{3N \times 3N}$ satisfying \par

% \begin{align}
% \begin{bmatrix}
% \Pi(\tilde{\mathbf{R}}_1)\mathbf{U}\! \mathbf{\Sigma}^{1 / 2}\mathbf{Y}_{1,.}^\transpose, \cdots, \Pi(\tilde{\mathbf{R}}_F) \mathbf{U} \mathbf{\Sigma}^{1 / 2}\mathbf{Y}_{N,.}^\transpose
% \end{bmatrix}
% \!=\!\mathbf{U}'\! \mathbf{\Sigma}^{1 / 2}\!\mathbf{Y}^\transpose.	
% \end{align}

% Following the remark \ref{remark_infinite_orthogonal}, when the rest matrices are given, there are infinite $\mathbf{U}'$ if $\frac{3N(3N-1)}{2} > F$. Therefore, there are infinite matrix $\tilde{\mathbf{X}}=\mathbf{U}' \mathbf{\Sigma}^{1 / 2}$ exists on condition $\frac{3N(3N-1)}{2} > F$.
% \end{proof}

We next take advantage of the zero derivatives and propagate the noise $\mathbf{E}$ to $\mathbf{X}^\mathrm{d}$ and $\mathbf{Y}^\mathrm{d}$. Denote $\mathbf{Q}$ as the 0-1-valued `row selection matrix' to select components of $\tilde{\mathbf{W}}$. $\lVert\mathbf{W}\lVert_F^2$ and $\lVert\mathbf{Q} \circ \tilde{\mathbf{W}}\lVert_F^2$ are equivalent ($\circ$ is the Hadamard product). Following remark \ref{lemma_infiite_solu} and by reaching the optimal solution, the nuclear norm of the objective function \eqref{Eq_NRSfM_objective} is known and there are exact or approximate $\dot{\mathbf{U}}_{\Delta}$ satisfying

\begin{equation}
\label{Eq_UdeltaXY_W}
% \begin{aligned}
\begin{split}
&f=\frac{1}{2}\lVert\mathbf{R}\mathbf{S}-\mathbf{W}\lVert_F^2=\frac{1}{2}\lVert\mathbf{Q} \circ (\mathbf{S}-\tilde{\mathbf{W}})\lVert_F^2\\
&=\frac{1}{2}\lVert\mathbf{Q}^\sharp \circ (\tilde{\mathbf{S}}^\sharp-\tilde{\mathbf{W}}^\sharp)\lVert_F^2
%&=\frac{1}{2}\lVert\mathbf{Q}^\sharp \circ (\tilde{\mathbf{X}}\mathbf{Y}^\transpose-\tilde{\mathbf{W}}^\sharp)\lVert_F^2\\
\!=\!\frac{1}{2}\lVert\dot{\mathbf{U}}_{\Delta}\hat{\mathbf{X}}\hat{\mathbf{Y}}^\transpose\!-\!g(\mathbf{W})\lVert_F^2\
\end{split},
\end{equation}

\noindent where $\mathbf{Q}^\sharp = g(\mathbf{Q})$ and $\tilde{\mathbf{W}} \in \mathbb{R}^{3F \times N}$ is the 3D version of $\mathbf{W}$ with zero values on Z direction. Note that first $2N$ rows of $\mathbf{Q}^\sharp$ ($3N \times F$) are all 1 and last $N$ rows are 0.\par

With regard to the objective function \eqref{Eq_UdeltaXY_W}, the optimized shrinkage optimizer $\mathbf{X}^{\mathrm{d}}$ and $\mathbf{Y}^{\mathrm{d}}$ should satisfy the local minima of \eqref{Eq_UdeltaXY_W} denoted as $f(\mathbf{X}^{\mathrm{d}},\mathbf{Y}^{\mathrm{d}})$. $f(\mathbf{X}^{\mathrm{d}},\mathbf{Y}^{\mathrm{d}})$ is close to noise-free estimators $f(\hat{\mathbf{X}},\hat{\mathbf{Y}})$ meaning the first-order expansion is close to the zero matrix defined as $\mathcal{O}$ (arbitrary size). For simplicity, we denote $\dot{\mathbf{U}}_{\Delta}\hat{\mathbf{X}}$ as $\dot{\hat{\mathbf{X}}}$ and $\dot{\mathbf{U}}_{\Delta}\mathbf{X}^\mathrm{d}$ as $\dot{\mathbf{X}}^{\mathrm{d}}$. $\dot{\mathbf{1}}$ is the first $2N$ rows of $\mathbf{1}$.\par

\begin{equation}
\label{Eq_provePartialZero1}
\begin{split}
% \begin{aligned}
&\frac{\partial{f(\mathbf{X}^{\mathrm{d}},\mathbf{Y}^{\mathrm{d}})}}{\partial{\mathbf{X}^{\mathrm{d}}}}
\stackrel{(\mathrm{i})}= \dot{\mathbf{ U}}_{\Delta}^\transpose(\dot{\mathbf{ U}}_{\Delta}\mathbf{X}^{\mathrm{d}}\mathbf{Y}^{\mathrm{d}\transpose}-g(\mathbf{W}))\mathbf{Y}^{\mathrm{d}}\\
&=[\dot{\mathbf{ U}}_{\Delta}^\transpose(\dot{\mathbf{ U}}_{\Delta}\mathbf{X}^{\mathrm{d}}{\mathbf{Y}^{\mathrm{d}{\transpose}}}-\dot{\mathbf{ U}}_{\Delta}\hat{\mathbf{X}} {\hat{\mathbf{Y}}^{\transpose}}-\mathbf{E}^\sharp)]\mathbf{Y}^\mathrm{d}\\
&\stackrel{(\mathrm{ii})}=(\dot{\mathbf{ U}}_{\Delta}^\transpose\dot{\mathbf{ U}}_{\Delta}\mathbf{X}^{\mathrm{d}}{\mathbf{Y}^{\mathrm{d}\transpose}}-\dot{\mathbf{ U}}_{\Delta}^\transpose\dot{\mathbf{ U}}_{\Delta}\hat{\mathbf{X}} \mathbf{H}^\transpose{\mathbf{Y}^{\mathrm{d}\transpose}}-\\
&\ \dot{\mathbf{ U}}_{\Delta}^\transpose\mathbf{E}^\sharp+\dot{\mathbf{ U}}_{\Delta}^\transpose\dot{\mathbf{ U}}_{\Delta}\mathbf{G}_2^\sharp))\mathbf{Y}^\mathrm{d}
=\mathcal{O},
\end{split}
\end{equation}

\vspace{-0.2cm}
\begin{equation}
\label{Eq_provePartialZero2}
\begin{split}
&\frac{\partial{f(\mathbf{X}^{\mathrm{d}},\mathbf{Y}^{\mathrm{d}})}}{\partial{\mathbf{Y}^{\mathrm{d}}}}
\stackrel{(\mathrm{i})}=\dot{\mathbf{X}}^{\mathrm{d}\transpose}(\dot{\mathbf{X}}^{\mathrm{d}}\mathbf{Y}^{\mathrm{d}\transpose}-g(\mathbf{W}))\\
&=\dot{\mathbf{X}}^{\mathrm{d}\transpose}(\dot{\mathbf{X}}^{\mathrm{d}}\mathbf{Y}^{\mathrm{d}\transpose}-\dot{\hat{\mathbf{X}}} {\hat{\mathbf{Y}}^{\transpose}}-\mathbf{E}^\sharp)\\
&\stackrel{(\mathrm{ii})}=\dot{\mathbf{X}}^{\mathrm{d}\transpose}(\dot{\mathbf{X}}^{\mathrm{d}}{\mathbf{Y}^{\mathrm{d}\transpose}}-\dot{\mathbf{X}}^{\mathrm{d}}\mathbf{H} {\hat{\mathbf{Y}}^{\transpose}}-(\mathbf{E}^\sharp+\dot{\mathbf{ U}}_{\Delta}{\mathbf{G}}_1^{\sharp}))=\mathcal{O},
\end{split}
\end{equation}

% \begin{align}
% % \begin{aligned}
% &\frac{\partial{f_i(\mathbf{X}^{\mathrm{d}},\mathbf{Y}^{\mathrm{d}})}}{\partial{\mathbf{Y}^{\mathrm{d}}}}
% \stackrel{(\mathrm{i})}=\dot{\mathbf{X}}_i^{'\transpose}(\dot{\mathbf{X}}'_i\mathbf{Y}^{\mathrm{d}\transpose}-\mathbf{W}^{\sharp})\\
% &=\dot{\mathbf{X}}_i^{'\transpose}(\dot{\mathbf{X}}'_i\mathbf{Y}^{\mathrm{d}\transpose}-\dot{\hat{\mathbf{X}}}_i^{'} {\hat{\mathbf{Y}}^{\transpose}}-\dot{\mathbf{E}})\\
% \label{Eq_provePartialZero2}
% &\stackrel{(\mathrm{ii})}\approx\dot{\mathbf{X}}_i^{'\transpose}(\dot{\mathbf{X}}'_i{\mathbf{Y}^{\mathrm{d}\transpose}}-\dot{\mathbf{X}}'_i\mathbf{H} {\hat{\mathbf{Y}}^{\transpose}}-\dot{\mathbf{E}})\approx\mathcal{O},
% \end{align}

\vspace{-0.1cm}
\noindent where (i) is from \eqref{Eq_UdeltaXY_W} and (ii) is from remark \ref{remark_upperbound}. The same conclusion applies. The noise of $\mathbf{W}^\sharp$ is $\mathbf{E}^\sharp=g(\mathbf{E})$ and follows $\mathcal{N}(0, \sigma_0^{2})$. We manipulate \eqref{Eq_provePartialZero1} and \eqref{Eq_provePartialZero2} to 

\vspace{-0.2cm}
\begin{equation}
% \begin{aligned}
\label{Eq_1st_proof}
\begin{split}
&\dot{\mathbf{ U}}_{\Delta}^\transpose\dot{\mathbf{ U}}_{\Delta}(\mathbf{X}^{\mathrm{d}}-\hat{\mathbf{X}}\mathbf{H}^\transpose){\mathbf{Y}^{\mathrm{d}\transpose}}\mathbf{Y}^{\mathrm{d}}= (\dot{\mathbf{ U}}_{\Delta}^\transpose\mathbf{E}^\sharp+\dot{\mathbf{ U}}_{\Delta}^\transpose\dot{\mathbf{ U}}_{\Delta}\mathbf{G}_2^{\sharp})\mathbf{Y}^{\mathrm{d}}\\
&\mathcal{E}(\dot{\mathbf{ U}}_{\Delta}^\transpose\dot{\mathbf{ U}}_{\Delta})(\mathbf{X}^{\mathrm{d}}\mathbf{H}\!-\!\hat{\mathbf{X}})\!\stackrel{(\mathrm{i})}=\\
&(\mathcal{E}(\dot{\mathbf{ U}}_{\Delta}^\transpose)\mathbf{E}^\sharp+\mathcal{E}(\dot{\mathbf{ U}}_{\Delta}^\transpose\dot{\mathbf{ U}}_{\Delta})\mathbf{G}_2^{\sharp})\mathbf{Y}^{\mathrm{d}}\!({\mathbf{Y}^{\mathrm{d}\transpose}}\mathbf{Y}^{\mathrm{d}})^{-1}\!\mathbf{H}\\
&(\mathbf{X}^{\mathrm{d}}\mathbf{H}\!-\!\hat{\mathbf{X}})^\transpose\!\mathrm{p}\mathbf{I}_{3N}\mathbf{e}_j\! \stackrel{(\mathrm{ii})}\approx\! \left[\frac{1}{\sqrt{3N}}\dot{\mathbf{1}}\mathbf{E}^\sharp\mathbf{Y}^{\mathrm{d}}({\mathbf{Y}^{\mathrm{d}\transpose}}\!\mathbf{Y}^{\mathrm{d}})^{-1}\!\mathbf{H}\right]^\transpose\!\mathbf{e}_j\\
&\stackrel{(\mathrm{iii})}{\sim} \mathcal{N}(0, \sigma_0^{2}(\mathbf{\Sigma}^{\mathrm{d}})^{-1})\\
&(\mathbf{X}^{\mathrm{d}}\mathbf{H}-\hat{\mathbf{X}})^\transpose\mathbf{e}_j
{\sim} \mathcal{N}(0, \frac{\sigma_0^{2}}{\mathrm{p}}(\mathbf{\Sigma}^{\mathrm{d}})^{-1}).
\end{split}
\end{equation}

\vspace{-0.4cm}
\begin{equation}
\label{Eq_2st_proof}
\begin{split}
&\dot{\hat{\mathbf{X}}}^{\transpose}\dot{\hat{\mathbf{X}}}({\mathbf{Y}^{\mathrm{d}\transpose}}-\mathbf{H}{\hat{\mathbf{Y}}^{\transpose}})= \dot{\hat{\mathbf{X}}}^{\transpose}(\mathbf{E}^\sharp+\dot{\mathbf{ U}}_{\Delta}{\mathbf{G}}_1^{\sharp})\\
&\mathcal{E}(\dot{\hat{\mathbf{X}}}^{\transpose}\dot{\hat{\mathbf{X}}})(\mathbf{H}^\transpose{\mathbf{Y}^{\mathrm{d}\transpose}}-{\hat{\mathbf{Y}}^{\transpose}})\stackrel{(\mathrm{i})}= \mathcal{E}(\dot{\hat{\mathbf{X}}}^{\transpose})(\mathbf{E}^\sharp+\dot{\mathbf{ U}}_{\Delta}{\mathbf{G}}_1^{\sharp})\\
&\mathrm{p}\mathbf{\Sigma}^{\mathrm{d}}(\mathbf{H}^\transpose{\mathbf{Y}^{\mathrm{d}\transpose}}-{\hat{\mathbf{Y}}^{\transpose}})\mathbf{e}_j\\
&\stackrel{(\mathrm{ii})}\approx \frac{1}{\sqrt{3N}}\dot{\mathbf{1}}\mathbf{\Sigma}^{\mathrm{d} 1 / 2}{\mathbf{E}^\sharp}
\stackrel{(\mathrm{iii})}{\sim} \mathcal{N}(0, {\sigma_0^{2}}(\mathbf{\Sigma}^{\mathrm{d}}))\\
&(\mathbf{H}^\transpose{\mathbf{Y}^{\mathrm{d}\transpose}}-{\hat{\mathbf{Y}}^{\transpose}})\mathbf{e}_j
{\sim} \mathcal{N}(0, \frac{\sigma_0^{2}}{\mathrm{p}}(\mathbf{\Sigma}^{\mathrm{d}})^{-1}).
\end{split}
\end{equation}

\vspace{-0.1cm}
Equation \eqref{Eq_1st_proof} and \eqref{Eq_2st_proof} only consider the Gaussian noises. (i): The Stiefel products $\dot{\mathbf{ U}}_{\Delta}^\transpose\dot{\mathbf{ U}}_{\Delta}$ and $\dot{\hat{\mathbf{X}}}^{\transpose}\dot{\hat{\mathbf{X}}}$ matrix are low-rank and not invertible; we approximate with their expectant following \cite{chen2019inference} due to the large number of frames. We use the expectant of all $F$ $\dot{\mathbf{ U}}_{\Delta}$ (in \eqref{Eq_1st_proof}) to enable the inverse operation. (ii): It follows the remark \ref{remark_U_expect} and ignores the non-Gaussian residuals (remark \ref{remark_upperbound}). 

Following remark \ref{remark_upperbound}, \eqref{Eq_1st_proof} and \eqref{Eq_2st_proof}, we can easily draw conclusion that the non-Gaussian errors are always smaller than the Gaussian noises. We provide the proof of (iii) below. Denote $\mathbf{E}'=\frac{1}{\sqrt{3N}}\mathbf{1}\mathbf{E}^\sharp$ (each element $\mathbf{E}'_{uv} \stackrel{i.i.d.}{\sim} \mathcal{N}(0, \frac{\sigma_0^{2}}{p}$).

\vspace{-0.1cm}
\begin{equation}
%\label{Eq_proof_1}
\begin{aligned}
&\left[\frac{1}{\sqrt{3N}}\mathbf{1}\mathbf{E}^\sharp\mathbf{Y}^{\mathrm{d}}({\mathbf{Y}^{\mathrm{d}\transpose}}\!\mathbf{Y}^{\mathrm{d}})^{-1}\!\mathbf{H}\right]^\transpose\!\mathbf{e}_j\!=\!\left[\mathbf{E}'\mathbf{Y}^{\mathrm{d}}\!({\mathbf{Y}^{\mathrm{d}\transpose}}\!\mathbf{Y}^{\mathrm{d}})^{-1}\right]\!^\transpose\!\mathbf{e}_j\\
&=\left[\sum_{i=1}^{F}\frac{1}{\lVert\mathbf{Y}^\mathrm{d}_{.,1}\lVert_2^2}{\mathbf{Y}}^\mathrm{d}_{i1}\mathbf{E}'_{vi} \ \cdots \  \sum_{i=1}^{F}\frac{1}{\lVert{\mathbf{Y}}^\mathrm{d}_{.,r}\lVert_2^2}{\mathbf{Y}}^\mathrm{d}_{ir}\mathbf{E}'_{vi}\right].
\end{aligned}
\end{equation}

Take the first element $\sum_{i=1}^{F}\frac{1}{\lVert\mathbf{Y}^\mathrm{d}_{.,1}\lVert_2^2}{\mathbf{Y}}^\mathrm{d}_{i1}\mathbf{E}'_{vi}$ as an example

\vspace{-0.1cm}
\begin{align}
\begin{split}
&\operatorname{Var}(\sum_{i=1}^{F}\frac{1}{\lVert\mathbf{Y}^\mathrm{d}_{.,1}\lVert_2^2}{\mathbf{Y}}^\mathrm{d}_{i1}\mathbf{E}'_{vi})=(\sum_{i=1}^{F}\frac{1}{\lVert\mathbf{Y}^\mathrm{d}_{.,1}\lVert_2^2}{\mathbf{Y}}^\mathrm{d}_{i1})^2\frac{\sigma_0^2}{p}=\\
&\frac{1}{(\lVert\mathbf{Y}^\mathrm{d}_{.,1}\lVert_2^2)^2}(\sum_{i=1}^{F}{\mathbf{Y}}^\mathrm{d}_{i1})^2\frac{\sigma_0^2}{p}\!=\!\frac{1}{(\lVert\mathbf{Y}^\mathrm{d}_{.,1}\lVert_2^2)}\sigma_0^2\!=\!\frac{\sigma_0^2}{p}(\mathbf{\Sigma}^{\sharp\mathrm{d}}_{11})^{-1}.
\end{split}
\end{align}

% \begin{remark}
% \label{remark_expect_U2}
% The expects $\mathcal{E}\!(\dot{\mathbf{ U}}_{\Delta}^\transpose\!\dot{\mathbf{ U}}_{\Delta})\!=\!\mathrm{p}\!\mathbf{I}_{3N}$ and $\mathcal{E}\!(\dot{\mathbf{ U}}_{\Delta}^\transpose)\!=\!\frac{1}{\sqrt{3N}}\!\dot{\mathbf{1}}$. It is from the definition of the orthogonal matrix \cite{aitken2017determinants}.
% \end{remark}

%\addtolength{\textheight}{-12cm} 
{\balance
\bibliographystyle{IEEEtran}
\bibliography{bib/strings-abrv,bib/ieee-abrv,reference} 
}

\end{document}